\documentclass[10pt]{article} % For LaTeX2e

\usepackage[accepted]{rlj}

\usepackage{amssymb}            % Defines common symbols like \mathbb R
\usepackage{mathtools}          % Extends amsmath, providing common math tools
\usepackage{mathrsfs}           % Enables \mathscr, which can work in cases that \mathcal does not
\usepackage{graphicx}           % For including images
\usepackage{subcaption}         % Allows for the use of subfigures and subcaptions
\usepackage[space]{grffile}     % For spaces in image names
\usepackage{url}                % For displaying URLs
\usepackage{lipsum}             % For placeholder text

\usepackage{algorithm}

\usepackage{bm}
\usepackage{amssymb}
\usepackage{subcaption}
\usepackage{footmisc}
\usepackage{wrapfig}
\usepackage{utfsym}
\usepackage{dsfont}
\usepackage{amsmath}
\usepackage{amsthm}

\usepackage{graphicx}
\usepackage{booktabs}

\usepackage{algorithm}
\usepackage{algpseudocode}

\newtheorem{Definition}{Definition}

\title{Credit Assignment and Focused Exploration for Sparse-reward Multi-agent Deep Reinforcement Learning}

\setrunningtitle{Credit Assignment and Focused Exploration for Sparse-reward MADRL}

\author{Shuai Han, Mehdi Dastani, Shihan Wang}

\emails{\{s.han,m.m.dastani,s.wang2\}@uu.nl}

\affiliations{
\textbf{Utrecht University, Utrecht, the Netherland}\\
}

\contribution{
    We introduce Influence Scope of Agents (ISA), a framework that automatically identifies the state dimensions/variables influenced by each agent via mutual information estimation, providing an influence-based decomposition of individual goals in 
    sparse-reward MADRL.}
    {
    Prior goal-conditioned MADRL methods typically solve the problem by replying on predefined subtasks, learned latent embeddings, or task-specific structures to define agent objectives. In contrast, ISA discovers agent-specific goal structures based on the mutual information learned from online interaction, without requiring offline data, prior task decomposition, or dense reward signals.
    }
\contribution{
    Based on the learned influence scopes, we propose a goal decomposition and credit assignment mechanism among agents that distributes intrinsic rewards according to action's influence on the goal.
    }
    {
    Existing multi-agent credit assignment methods typically rely on centralized critics, value decomposition, or implicit network-based mechanisms. Different from these methods, ISA explicitly models the influence of agents on different parts of the state for credit assignment. 
    }

\contribution{
    We design a focused exploration strategy that restricts intrinsic count-based exploration to the state dimensions influenced by the agent, reducing the effective exploration space in sparse-reward multi-agent environments.
    }
    {
    Standard exploration methods in MADRL typically operate on full-state representations, while ISA leverages agent-specific controllable subspaces to guide exploration.
    }

\contribution{
    We empirically demonstrate that ISA achieves substantial performance gains in challenging sparse-reward benchmarks, outperforming state-of-the-art methods by up to 56\% on tasks where baselines are able to learn, and successfully solving several difficult scenarios where baseline methods fail to converge.
    }
    {
    None.
    }

\keywords{Multi-agent deep reinforcement learning, exploration, credit assignment.} % Your keywords

\summary{Recent advances in multi-agent deep reinforcement learning (MADRL) have achieved strong performance in various scenarios. However, training cooperative policies in sparse-reward scenarios remains a major challenge for MADRL due to the unfocused exploration and ambiguous credit assignment. In this paper, we introduce Influence Scope of Agents (ISA), an algorithm that leverages agents' influence for efficient policy training. By evaluating the mutual dependence between agents' actions and states, it automatically learns the scope of state dimensions (attributes) that can be influenced by individual agents. These influence scopes are then used to focus agents' exploration on their controllable aspects of the environment and to calculate credit assignment among agents according to their influence. We evaluate ISA in a variety of sparse-reward multi-agent scenarios. The results show that our method significantly outperforms state-of-the-art baselines.

}

\begin{document}

\makeCover  % Create the cover page
\maketitle  % Make the title section

\begin{abstract}
Recent advances in multi-agent deep reinforcement learning (MADRL) have achieved strong performance in various scenarios. However, training cooperative policies in sparse-reward scenarios remains a major challenge for MADRL due to the unfocused exploration and ambiguous credit assignment. In this paper, we introduce Influence Scope of Agents (ISA), an algorithm that leverages agents' influence for efficient policy training. By evaluating the mutual dependence between agents' actions and states, it automatically learns the scope of state dimensions (attributes) that can be influenced by individual agents. These influence scopes are then used to focus agents' exploration on their controllable aspects of the environment and to calculate credit assignment among agents according to their influence. We evaluate ISA in a variety of sparse-reward multi-agent scenarios. The results show that our method significantly outperforms state-of-the-art baselines\footnote{The code has been open-sourced at \url{https://github.com/shan0126/ISA/tree/main}
}.
\end{abstract}

%%%%%%%%%%%%%%%%%%%%%%%%%%%%%%%%%%%%%%%%%%%%%%%%%%%%%%%%%%%%%%%%
%% Section: Submission of papers to RLJ/RLC
%%%%%%%%%%%%%%%%%%%%%%%%%%%%%%%%%%%%%%%%%%%%%%%%%%%%%%%%%%%%%%%%

\section{Introduction}

Multi-Agent Deep Reinforcement Learning (MADRL) has been widely applied in various fields in recent years, such as autonomous driving \citep{apply_driving}, traffic signal control \citep{apply_traffic}, and unmanned aerial vehicles \citep{apply_uva}. However, the success of MADRL applications heavily relies on handcrafted reward functions to provide immediate feedback to agents. In sparse reward scenarios, MADRL methods show low sample efficiency \citep{CMAE} or even fail to learn \citep{laies}. Inspired by solutions for single-agent sparse-reward domains \citep{HER}, goal-conditioned MADRL has recently emerged as a promising approach by measuring individual goal achievement as intrinsic individual rewards for agents. With predefined subtasks \citep{ALMA}, sampled observations \citep{MASER} or latent variables \citep{HMASD} as goals, previous methods are able to learn competitive policies in sparse-reward MADRL tasks.

However, there are still some open challenges for goal-conditioned MADRL. 
Firstly, the multi-agent domain inherently amalgamates information from multiple agents \citep{MPE,SMAC}, which brings challenges to automatically delimit dimensions/attributes from environmental states for measuring individual goal achievement. 
Treating wrong information as an agent's goal can be harmful \citep{SurveyIntrinsically}.
For example, when training an agent to pick up an apple, it may not make sense to use the position of another agent as a goal.
Secondly, the individual goal achievement of an agent may be affected by other agents, which means an agent may receive non-stationarity feedback because of the actions from other agents \citep{laies}. 
Thirdly, the state and joint action spaces of MADRL increase exponentially with the number of agents \citep{HMASD}, which poses a challenge for exploring valuable states to identify the value of goals for specific tasks. 

% unstable or even wrong feedback

Aiming at solving the individual goal delimitation, credit assignment and exploration problems for goal-conditioned MADRL in sparse-reward tasks, we propose Influence Scope of Agents (ISA). ISA introduces the concept of influence scope for agents into multi-agent system, which can be efficiently and automatically calculated by measuring the mutual dependence between agents' discrete actions and state attributes/dimensions. This is done using the well-known information theoretic concept of mutual information between variables \citep{shannon1959coding}. Such influence scope of an agent delimits its individual goal space to provide succinct goal representation. Additionally, by identifying the joint influence scope of all agents, it can be automatically determined which segment of individual goal may be influenced by the team of agents. In credit assignment, the agent will not be rewarded from the segment if its current action cannot influence this segment, thus reducing the non-stationary evaluation on individual policies. Moreover, the influence scope is also used to downscale the individual exploration space by excluding the dimensions/attributes that cannot be influenced by this agent to improve exploration.

%hereby improving the efficiency of exploration.

We verify the performance of our method on multiple tasks of challenging multi-agent sparse-reward environments. ISA outperforms state-of-the-art methods by up to 56\% on tasks where baseline methods are able to learn, and uniquely succeeds in several challenging tasks where baselines fail entirely. Ablation experiments demonstrate the effectiveness of the proposed credit assignment and exploration methods based on influence scope. We also show the interpretability of ISA on the credit assignment among agents during training.

\section{Related Work}

\textbf{Factored Dec-POMDPs.} In Decentralized Partially Observable Markov Decision Processes (Dec-POMDPs), considering subsets of state variables, instead of the full state, can improve the scalability of algorithms. The assumption that the global state can be decomposed into conditionally independent subsets of variables \citep{oliehoek2008exploiting} enables compact representations of transition and observation functions. This structure supports more efficient planning \citep{oliehoek2008exploiting, allen2009complexity} and scales better with increasing agent numbers \citep{pajarinen2011efficient, oliehoek2013approximate}. Our method draws inspiration from this idea. However, we do not rely on this formalism as we do not introduce conditional independence assumptions. Instead, all actions may influence all state variables, with some having stronger effects on specific variables. Moreover, unlike prior work, our influence structure is not predefined but learned from interaction.

\textbf{Credit assignment in sparse-reward domains.} In MADRL, credit assignment is typically achieved by the mixing value function \citep{QMIX,QTRAN} or a centralized critic \citep{DOP,COMA} to decompose the team reward. In the sparse-reward environments where value functions and critics fail to learn, goal-conditioned MADRL\citep{CM3, MASER} introduce individual goals to provide intrinsic reward. Building on these methods, this work further studies better representation of individual goals in MADRL and introduces a novel credit assignment method among agents based on their influences. Unlike methods that rely on prior knowledge \citep{laies} or redistribute delayed global rewards through auxiliary attention \citep{delayr1, delayr2, delayr3, she2022agent}, our method consider a fully sparse-reward setting where no prior knowledge or delayed rewards are available. ISA discovers goals and learns purely through online interaction, without requiring offline data, prior task structure, or dense feedback signals.

% studies the representation of individual goals in MARL, proposes the individual goal representation based on influence scope of agent, and introduces a novel credit assignment method among agents based on their influences.

\textbf{Information-theoretic exploration or coordination.} Information theoretic methods are often used to quantify learning signals \citep{ROMA,PMIC}, especially in sparse-rewards domains. Empowerment-based methods \citep{salge2014changing,dai2023empowerment} and CMAE \citep{CMAE} use information theory to enhance exploration. EITI \citep{EITI} quantify influence of one agent's behavior on the reward. LAIES \citep{laies} tracks lazy agent by investigating causality. FoX \citep{FoX} quantifies formations in MADRL via mutual information. HMASD \citep{HMASD} coordinate different agents by maximizing the mutual information between state and skills.
Unlike these works, the information theory in our work is used to automatically determine the influence scope of agents on the state for credit assignment and exploration. A detailed related work can be found in Appendix A.

\section{Preliminaries}

\textbf{Dec-POMDP.} In line with previous goal-conditioned MADRL \citep{MASER}, our method is formulated based on the Dec-POMDP \citep{rebuttal_decpomdp1, DecPOMDPs}, which is defined as a tuple $G = <\mathcal{I}, \mathcal{S}, \mathcal{A}, \mathcal{P}, O, \Omega, R, \gamma>$, where $\mathcal{I}$ is the set of $N$ agents, $\mathcal{S}$ is the global state space of the environment, $\mathcal{A} = \mathcal{A}_1 \times \mathcal{A}_2 \times \dots \times\mathcal{A}_N$ is the joint action space and $\mathcal{A}_i$ is the action space of an individual agent $i$, $\mathcal{P}: \mathcal{S} \times \mathcal{A} \times \mathcal{S} \rightarrow [0, 1]$ is the transition probability function, $\Omega$ is the observation space, $O: \mathcal{S} \times \mathcal{I} \rightarrow \Omega$ is the observation function, $R$ is the shared reward function, and $\gamma \in [0, 1)$ is a discounted factor. When interacting with the environment, each agent $i$ draws observation $o_i \in O(s,i)$, where $s\in \mathcal{S}$ denotes the current global state. Then, each agent $i$ samples its action $a_i \in \mathcal{A}_i$ with a stochastic policy $\pi_i: \mathcal{T}_i \times \mathcal{A}_i \rightarrow [0, 1]$ where $\mathcal{T}_i = (\Omega \times \mathcal{A}_i)^* \times \Omega$ represents the trajectory of agent $i$ where $(\Omega \times \mathcal{A}_i)^*$ represents the Kleene closure on $\Omega \times \mathcal{A}_i$. After executing the joint action $ \mathbf{a} = [a_1, ..., a_N]$, the system transitions to a next state $s' \in \mathcal{S}$ and receives a shared reward $r$ from $R$. The target of fully cooperative MADRL is to learn the team policy to maximize the excepted accumulated reward. In this work, we particularly consider the sparse-reward setting where the nonzero reward is not given to agents’ actions in every step but only when certain conditions are met \citep{HMASD,FoX}. Our methods follow the centralized training \& decentralized execution (CTDE) paradigm \citep{rebuttal_ctde1, QMIX} of MADRL where global information is available in training, but in execution, only local information is available. 

% Mutual information is widely used in MARL to provide quantifiable metrics to assist in policy learning.

% given the probability distributions $p(x), p(y)$ and the joint probability distributions $p(x, y)$ of two variables $X$ and $Y$ 

% Formally, the mutual information between two variables $X$ and $Y$ is:

\textbf{Mutual information.} Mutual information quantifies dependence between two variables, which is widely used in MADRL to assist in policy learning. Given the probability distributions $p(x), p(y)$ and the joint probability distributions $p(x, y)$ of two variables $X$ and $Y$, the mutual information $I(X;Y) = \sum_{x,y} p(x, y) \log \frac{p(x)p(y)}{p(x,y)} = H(X) - H(X|Y)$,
where $H(\cdot)$ and $H(\cdot|\cdot)$ represent the entropy and conditional entropy respectively, and $x$ and $y$ are values from the range of variables $X$ and $Y$, respectively. This equation can be read as how much knowing $Y$ reduces uncertainty about $X$. 
The conditional mutual information is introduced when the information gain between two variables $X$ and $Y$ is conditioned upon a third variable $Z$, which can be calculated by $I(X;Y|Z) = \mathbb{E}_{z} [I(X; Y | Z = z)]$, where $z$ represents a certain value of $Z$.

\textbf{Additional notations.} We now introduce some specific notations in this work. Given a Dec-POMDP, the state $s\in\mathcal{S}$ can be represented as a $K$-dimensional vector $s = [s^1, s^2, ..., s^K]$, where $s^k \in \mathbb{R}$ denotes the value on $k$-th dimension of state $s$. Given a state vector $s$ and an index set $D \subseteq \{1, 2,..., K\}$, we use Proj$_{D}(s)=(s^k)_{k\in D}$ to restrict the state vector to attributes indexed by $D$. For instance, for $D = \{3, 7\}$, Proj$_{D}(s)=(s^3, s^7)$ takes the attribute values of $s$ at indexes $\{3, 7\}$.

\section{Core Concepts}

In this section, we introduce the concept \textit{influence scope}, based on the assumption that in a Dec-POMDP, an action affects certain dimensions of the state more significantly then others. This reflects many real-world settings \citep{apply_driving, EITI}. For instance, in autonomous driving, gas pedal action of a vehicle significantly affects the data from speed sensor. In robotics tasks, a robot’s `walking' action has little impact on the positions of other robots.

Due to the nonlinear, high-dimensional, and entangled state dynamics across agents in MADRL, it is challenging to directly measure changes across the entire state vector. Therefore, we begin by analyzing how individual state dimensions are affected by an agent’s action, rather than attempting to model the entire state transition at once. Specifically, 
conditioned on the actions of other agents $\bm{a_{-i}}$, we measure the mutual dependence between the execution of action $a_i$ and the state change $\Delta s^k$ on dimension $k$ ($\Delta s^k = s'^k - s^k$ where $s^k$ and $s'^k$ represent the $k$ dimension of current and next state, respectively), which is denoted as $I(\Delta s^k;a_i|\bm{a_{-i}})$. This measures how much knowing the execution of action $a_i$ reduces uncertainty about the state change on $k$-th dimension given the action of other agents $\bm{a_{-i}}$. We use the value of this mutual information to quantify the influence of a certain action $a_i$ on $k$-th dimension of state. If this influence exceeds a certain threshold $\delta$, then we say the information on $k$-th dimension of state is influenced by this action $a_i$. In our method, $I(\Delta s^k;a_i|\bm{a_{-i}})$ is calculated from the expectation as follows.
\begin{equation}\label{Equ:I}
	\begin{split}
		I(\Delta s^k;a_i|\bm{a_{-i}}) =  \mathbb{E}_{\bm{\tilde{a}_{-i}}} [I(\Delta s^k;a_i | \bm{a_{-i}} = \bm{\tilde{a}_{-i}} )]
	\end{split}
\end{equation}
where $\bm{\tilde{a}_{-i}}$ represents specific values of variable $\bm{a_{-i}}$. When computing Equ. (\ref{Equ:I}), we need to calculate the mutual information under different specific values $\bm{\tilde{a}_{-i}}$ of $\bm{a_{-i}}$ and then take the expectation. If agents take random actions to collect interaction data, it will lead to an exponential variety of combinations for $\bm{a_{-i}}$ and thus may cause the calculation in Equ. (\ref{Equ:I}) to be intractable. In our practice, to estimate Equ. (\ref{Equ:I}) for agent $i$, we use Monte Carlo sampling \citep{I04}. Specifically, we randomly sample multiple combinations for $\bm{a_{-i}}$ to calculate the average and to estimate the expected value. As the sample size grows large, this average provides an unbiased estimate of the expected value. To estimate the distributions of $\Delta s^k$ and $a_i$, we discretize the continuous variable $\Delta s^k$ with equal width binning \citep{I04} and convert $a_i$ to a binary value: 1 if the current action is $a_i$, 0 otherwise. Note that the mutual information in Equ. (\ref{Equ:I}) is computed separately for each state dimension. As a result, the complexity grows linearly with the state dimension $K$, rather than exponentially with the joint state space. This design avoids high-dimensional density estimation and ensures computational scalability.

\begin{wrapfigure}{r}{0.28\linewidth}
    \vspace{-6mm}
    \centering
    
    \begin{subfigure}{1\linewidth}
        \includegraphics[width=\linewidth]{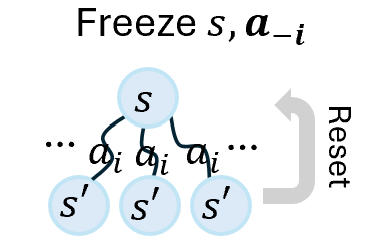}
        \caption{}
        \label{reset}
    \end{subfigure}
    
    \vspace{2mm}
    
    \begin{subfigure}{1\linewidth}
        \includegraphics[width=\linewidth]{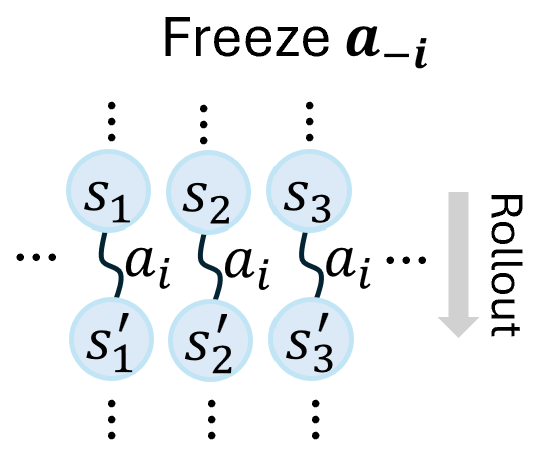}
        \caption{}
        \label{rollout}
    \end{subfigure}
    
    \vspace{-3mm}
    \caption{Data collection for different measurement}
    \vspace{-6mm}
\end{wrapfigure}

Notably, Equ. \ref{Equ:I} is to capture the general action influence in an environment. Prior work typically measure mutual information between variables in Dec-POMDPs conditioned on specific environment states \citep{EITI, dai2023empowerment}. Following this idea, the data for measuring mutual information need to be collected by freezing some specific states (as shown in Fig. \ref{reset}), which requires repeated resets to those states via a simulator. In contrast to this design, our method considers the general influence of an agent’s action across a broad distribution of encountered states. This allows the environment state to vary during random rollouts (as shown in Fig. \ref{rollout}). Therefore, our approach avoids the probably impractical need for environment resets.

After measuring the influence of an action $a_i$ of agent $i$ on specific dimension of states by $I(\Delta s^k;a_i|\bm{a_{-i}})$, we can define the influence scope of $a_i$ as follows.

\begin{Definition}[Influence scope of action]
	\label{Def:inf_ac}
	Given a Dec-POMDP, the influence scope $D(a_i)$ for action $a_i$ of agent $i$ is an index set including all the dimension indices of the state that are influenced by this action, which is denoted by: $D(a_i) = \{ k | I(\Delta s^k;a_i|\bm{a_{-i}}) > \delta, s,s'\in\mathcal{S} \} $, where $\delta \ge 0$ is a threshold.
\end{Definition}

\textbf{Remark 1 (Threshold-based influence scope of action):} In our method, threshold $\delta$ is a hyperparameter to examine the level/degree of influences of an agent's action on state dimensions. When $\delta=0$, all dimensions of the state will be recognized as being affected by all actions. Conversely, when $\delta$ is set to a very large value, the algorithm will perceive the influence scope of all actions as an empty set. Actually, which dimensions of the state are significantly affected by a specific action is an inherent property of the environment. In practice, we find suitable $\delta$ by fine-tuning. 

\textbf{Remark 2 (Credit assignment based on influence):}  The basic idea underlying Definition \ref{Def:inf_ac} is to use the influence scope to assign credits in multi-agent tasks. If a reward is caused by dimension $k$ of state and $k\notin D(a_i)$, then $a_i$ will not result in any reward. For example, by setting appropriate $\delta$, agents can be aware that their `walking' actions can significantly affect their location changes and their `pressing' actions significantly affect whether a button is pressed. In this case, if the team of agents receive a reward because the button is pressed at some time step, then the agent performing `pressing' will be assigned with this team reward and agents performing `walking' will not be.

% In MARL, the basic unit for receiving rewards and learning policies is the agent. Therefore, we define the influence scope of an agent based on the combined influence of all its actions.

Given the influence scope of an action, the agent's influence scope is introduced as follows.

\begin{Definition}[Influence scope of agent]
	\label{Def:inf_ag}
	Given a Dec-POMDP and influence scope of actions $D(a_i)$ for $ a_i\in A_i$, the influence scope of agent $i$ is an index set including all the dimension indices of the state that are influenced (for the given $\delta$) by the actions performed by this agent, which is denoted by: $D_{i} =  \cup_{a_i\in A_i} D(a_i)$.
	% \begin{equation}\label{Equ:Di_t}
		% 	D_{i} =  \cup_{l=1}^{L_i} D^{l}_i
		% \end{equation}
\end{Definition}

Definition 2 provides the influence scope of an agent through an index set of state dimensions affected by its actions\footnote{A special case arises when the team contains a dummy agent whose actions do not influence the states. In this case, its learned influence scope $D_i$ is empty. ISA does not explicitly handle this case. However, the learned influence scope provides a useful signal for identifying such agents, since an agent with an empty influence scope can be interpreted as having no detectable effect on the environment. To avoid this degenerate case and keep the subsequent formulation focused on agents that can affect the environment, we assume in the remainder of the paper that $D_i$ is non-empty for every agent and do not explicitly consider dummy agents.}. In MADRL, a desired state inherently reflects the combined influence of multiple agents. Individual influence scope in Definition 2 enables a compact, influence-based representation of individual goals.

\begin{Definition}[Global goal and individual goal]
	\label{Def:Goal_t}
	Given a Dec-POMDP and the influence scope $D_i$ of agent $i$, a global goal $g = [s^1,...,s^K] \in \mathcal{S}$ is a vector from the $K$-dimensional state space. Given a global goal $g \in \mathcal{S}$, an individual goal $g_i$ for an agent $i$ is a projection Proj$_{D_i}(g) = (s^k)_{k \in D_i}$ that takes the value of the input vector g only at the indices given by $D_i$. 
\end{Definition}

The goal representation in reinforcement learning is closely tied to the intrinsic reward computation that measures the goal achievement \citep{SurveyIntrinsically}. Following Definition 3, we propose to take the values of state attributes given by $D_i$ as agent $i$’s goal to provide intrinsic reward, which can be more efficient to stimulate agent $i$’s learning. 
This design is aligned with the concept of stimulus and reward in biology, where stimulus refers to environmental changes or signals that influence the actions of a living organism \citep{berridge2003parsing}. Note that $D_i$ denotes the dimensions of the environment that have mutual dependence with (and influenced by) the actions of agent $i$. In our method, global goals are successful terminal states that satisfy the task-success condition, which are revealed by the reward function in sparse-reward settings \citep{HMASD}. Therefore, the number of global goals depends on the set of successful terminal states rather than on the sparsity or density of intermediate rewards. Even when the environment provides dense or small intermediate rewards, ISA only stores successful terminal states as global goals.

% global goals are defined as the terminal states associated with task success, which are revealed by the reward function in sparse-reward settings \citep{HMASD}.

\textbf{Remark 3: (Non-conflicting individual goals):} When the influence scopes of different agents overlap, their individual goals based on influence scope may in principle conflict in terms of the overlapping part. However, since all individual goals are derived from the same global goal by Definition 3, the values on overlapping parts are always consistent. For example, if two agents jointly influence a switch variable with on/off values, their objective value on this variable wouldn't be contradictory because its value is uniquely defined from the same global goal. Importantly, since all individual goals are projections of the same global goal state, the target values on overlapping dimensions are inherently identical across agents. This guarantees value consistency and prevents objective misalignment during decentralized execution. As a result, ISA preserves cooperative coherence while allowing individualized goal representations.

\textbf{Remark 4: (Trainable environments):} Because the reward of a given Dec-POMDP depends on part or all the dimensions of state, there always exists an index set $D' \subseteq \{1, 2, ..., K\}$ where the values of these state dimensions determine the reward of the Dec-POMDP. The environment is trainable by ISA only if all dimensions in $D'$ can be influenced by the behavior of at least one agent, i.e., $D' \subseteq \cup_{i\in \mathcal{I}} D_i$. In fact, this condition can always be satisfied because, according to Remark 1, all dimensions will be considered to be within the agent's influence scope when $\delta = 0$. In this case, our approach is consistent with the classic approach of treating the entire state as the goal for agents \citep{SurveyIntrinsically}. And when the dimensions of state that determine the reward in the given Dec-POMDP are inherently influenced by the agents, the information gains underlying in Definition \ref{Def:inf_ac} $\sim$ \ref{Def:Goal_t} allows ISA to find the inherent influence scope in the given environment and the corresponding individual goals that determine the reward of Dec-POMDP.

Given individual goals, we define which segments are jointly influenced by the team and which are agent-specific.

\begin{Definition}[Common segment and special segment]
	\label{Def:segments}
	Given a Dec-POMDP, the influence scope $D_i$ for all $i\in\mathcal{I}$ and a global goal $g \in \mathcal{S}$, the common segment $g_i^c$ of agent $i$ is defined as the segment of its individual goal that affected by all agents, which is given by $g^c_i =$ Proj$_{D^c}(g)$, where $D^c = \cap_{i\in\mathcal{I}} D_{i}$. The special segment $g_i^{(i-c)}$ for agent $i$ is defined as the segment of its individual goal that excludes the common segment, which is given by $g_i^{(i-c)} =$ Proj$_{D^{(i-c)}}(g)$,where $D^{(i-c)} = D_{i} \setminus D^{c}$.
\end{Definition}
According to Definition \ref{Def:segments}, each agent has the same common segment on their individual goal, i.e., $g_i^c = g_j^c$ for all $i,j \in \mathcal{I}$. Besides, segments $g_i^c$ and $g_i^{(i-c)}$ on individual goal $g_i$ are given by index sets $D^c$ and $D^{(i-c)}$, which can be understood as joint influence scope of all agents and influence scope for agent $i$ without the joint influence.

\section{Algorithm}

Building on the concepts introduced above, a general process of the proposed ISA is as follows:

\begin{wrapfigure}{r}{0.5\linewidth}
    \centering
    \includegraphics[width=\linewidth]{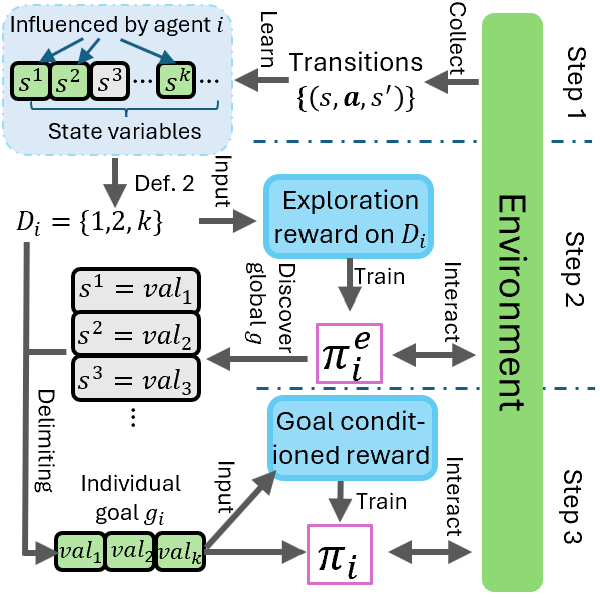}
    \vspace{-2mm}
    \caption{Three-step ISA workflow illustrated with example where agent $i$ influences dimensions $\{1,2,k\}$ of states.}
    \label{fig:frame}
    \vspace{-6mm}
\end{wrapfigure}

\textbf{Step 1. Obtain influence scope.} ISA first collects transitions by interacting with the environment and then computes the influence scopes $D(a_i)$ and $D_i$ for each $i \in \mathcal{I}$ and each $a_i \in A_i$ according to Definitions \ref{Def:inf_ac} and \ref{Def:inf_ag} based on these collected transitions. As shown by the example in Fig. 2, ISA learns from transitions that agent $i$ can influence variables $s^1, s^2$, and $s^k$, which means $D_i = \{1,2,k\}$.

\textbf{Step 2. Explore global goal.} With the influence scopes $D_i$, ISA trains the exploration policies $\{\pi^e_i\}_{i\in\mathcal{I}}$ where $\pi^e_i: \mathcal{T}_i \times \mathcal{A}_i \rightarrow [0, 1]$, to discover a set of success states as global goals. This training process is guided by exploration rewards defined with $D_i$ (which will be introduced in subsection \ref{sec:explora}). This process is illustrated in the middle of Fig. 2. Notably, ISA has no prior knowledge about the values of global goals. Therefore, it is necessary in this step to discover global goals.

\textbf{Step 3. Train goal-conditioned policies.} With at least one explored global goal $g$, ISA trains policies conditioned on individual goals $g_i$ decomposed from $g$ based on Definition \ref{Def:Goal_t}. Specifically, ISA trains goal-conditioned policies $\{\pi_i\}_{i\in\mathcal{I}}$ where $\pi_i: \mathcal{T}_i \times \mathcal{G}_i  \times \mathcal{A}_i \rightarrow [0, 1]$ and $\mathcal{G}_i$ represents the individual goal space. By uniformly sample a global goal $g$ among the discovered set of global goals for a whole episode, $g_i$, decomposed from $g$, is used to be a part of input of $\pi_i$ and to generate intrinsic rewards to train $\pi_i$. After repeated sampling of $g$ and sufficient training, agents can be trained towards achieving multiple global goals discovered in Step 2, which are guaranteed by the multi-goal reinforcement learning paradigm \citep{schaul2015universal,SurveyIntrinsically}. As shown in Fig. 2, since $D_i$ of agent $i$ is $\{1, 2, K\}$, the  individual goal derived from the global goal is $g_i = [val_1, val_2, val_k]$. Note that the dimensions influenced by agent $i$ are learned during Step 1, while the specific values ($val_1, val_2, ...$) of these dimensions on the global goal are explored in Step 2. The individual goal is input to the policy $\pi_i$ and for computing the goal-conditioned reward to train $\pi_i$. After centralized training, individual goals $\{g_i\}_{i\in\mathcal{I}}$, decomposed from a sampled global goal $g$, will be deployed locally to enable the decentralized execution of $\{\pi_i\}_{i\in\mathcal{I}}$.

% The rest of this section will explain in detail how to train the goal-conditioned policies $\{\pi_i\}_{i\in\mathcal{I}}$ and the exploration policies $\{\pi^e_i\}_{i\in\mathcal{I}}$.

% $\{\pi_i\}_{i\in\mathcal{I}}$ in Step 3 with goal-conditioned credit assignment, as well as how to train exploration policies $\{\pi^e_i\}_{i\in\mathcal{I}}$ in Step 2 with downscale individual exploration space.

\subsection{Goal-conditioned Credit Assignment}

To train the goal-conditioned policies $\{\pi_i\}_{i\in\mathcal{I}}$, previous work \citep{MASER, CM3} draw inspiration from measuring the goal achievement in single-agent domain \citep{SurveyCA}. However, in multi-agent scenarios where the influence of agents on the environment overlaps, the goal achievement of an agent may be influenced by others, which brings challenge to measure the agent's contribution of individual goal achievement. In order to address this problem, we first divide each individual goal into common and special segments according to Definition \ref{Def:segments}. When evaluating the behavior of an agent $i$, we measure the impact of this behavior on these two segments separately. The reward from common segment is:
\begin{equation}\label{Equ:RC}
	R^c_i(s, s' | g_i^c) = d(s^c, g_i^c) - d(s'^c, g_i^c)
\end{equation}
where $s^c =$ Proj$_{D^c}(s)$ and $s'^c =$ Proj$_{D^c}(s')$ are the restricted vectors of current and next states on dimensions given by $D^c$, $g_i^c$ is the common segment of individual goal given by Definition \ref{Def:segments} and $d$ is the distance metric function between two vectors. According to Equ. (\ref{Equ:RC}), this reward function produces a positive gain when the current state changes in a direction close to $g_i^c$. In this paper we use the combination of the Euclidean and Hamming distances: $d(v_1, v_2) = d_E(v_1, v_2) + \lambda d_H(v_1, v_2)$, where $v_1, v_2$ are the input vectors, $d_E$ is the Euclidean distance, $d_H$ is the Hamming distance, and $\lambda$ is the hyper-parameter factor. This combination is used because environment states often include continuous and discrete dimensions. Euclidean distance captures the former, while Hamming distance handles the latter. Similarly, the reward form special segment is:
%\begin{equation}\label{Equ:RS} R^{(i-c)}_i(s, a, s' | g_i^{(i-c)}) = d(\phi(s, D^{(i-c)}), g_i^{(i-c)}) - d(\phi(s', D^{(i-c)}), g_i^{(i-c)}) \end{equation}
\begin{equation}\label{Equ:RS}
	R^{(i-c)}_i(s, s' | g_i^{(i-c)}) = d(s^{(i-c)}, g_i^{(i-c)}) - d(s'^{(i-c)}, g_i^{(i-c)})
\end{equation}
where $s^{(i-c)} =$ Proj$_{D^{(i-c)}}(s)$ and $s'^{(i-c)} =$ Proj$_{D^{(i-c)}}(s')$ are the restricted vectors of current and next states on dimensions given by $D^{(i-c)}$, and $g_i^{(i-c)}$ is the special segment given by Definition \ref{Def:segments}.

With these rewards from two segments, we can more precisely assign credits to each agent based on the action influence to different segments. Specifically, given a transition $(s, \bm{a}, s')$ and a global goal $g$ sampled from discovered success states, each agent $i$'s goal-conditioned reward is:
\begin{equation} \label{Equ:GCR}
\begin{aligned}
R_i(s, a_i, s' \mid g_i) =
\left\{
\begin{array}{l}
R^c_i(s, s' \mid g_i^c) + \alpha_1 R^{(i-c)}_i(s, s' \mid g_i^{(i-c)}) \text{\;\;\;\ if } D(a_i) \cap D^c \neq \emptyset \\
\alpha_1 R^{(i-c)}_i(s, s' \mid g_i^{(i-c)})
\qquad \text{otherwise}
\end{array}
\right.
\end{aligned}
\end{equation}
where $g_i$ is the individual goal decomposed from the global goal $g$, and $\alpha_1$ is a factor to model the different importance of dimensions on the rewards. According to Equ. (\ref{Equ:GCR}), within $\bm{a}$, if the action $a_i$ of agent $i$ can affect the common segment of individual goal (i.e., $D_i(a) \cap D^c \neq \emptyset$), the goal-conditioned intrinsic individual reward of agent $i$ will be computed from both the common segment and the special segment. Otherwise, its intrinsic individual reward will only be computed from the special segment. In this way, the credit can be assigned among agents. Equ. (\ref{Equ:GCR}) distributes rewards based on the influence of actions. This design covers cases where actions affect all dimensions, and improves credit assignment when actions influence a subset of dimensions.

% Finally, goal-conditioned policies $\{\pi_i\}_{i\in\mathcal{I}}$ are trained with the combinations of intrinsic and environmental rewards: $r_i + \alpha_2 r$, where $r_i = R_i(s, a_i, s' | g_i)$, $r$ is the environmental reward and $\alpha_2$ is a scaling factor.

\subsection{Focused Exploration based on Influence Scope}
\label{sec:explora}

To train the exploration policies $\{\pi^e_i\}_{i\in\mathcal{I}}$ for discovering the success states as global goal, we draw inspiration from the counting-based exploration \citep{count1,count2} which assigns bonus rewards to encourage exploring new states. However, in multi-agent domain, counting the full state is ineffective due to the high-dimensionality \citep{count2}. To address this problem, ISA counts only the segments of the state that individual agents can influence, enabling more focused and efficient exploration. In our methods, the reward used for encouraging towards common segment jointly influenced by all agents is calculated by:$R_{i+}^c(s') = 1 / \sqrt{N(\varphi(s'^c))}$,
where $N(\cdot)$ returns the count of input vectors and $\varphi$ is a hash function, which maps states into a smaller set of bins to allow efficient visitation estimation. Every time a specific $s'^c =$ Proj$_{D^c}(s'^c)$ is encountered in the multi-agent system, $N(\varphi(s'^c))$ is increased by one. Similarly, the reward used for encouraging an individual agent towards the special segment solely influenced by itself is calculated by: $R_{i+}^{(i-c)}(s') = 1 / \sqrt{N(\varphi(s'^{(i-c)}))}$.
$R_{i+}^c(s')$ and $R_{i+}^{(i-c)}(s')$ measure the novelty of the state restricted by influence scopes. The more novel the projection, the greater the bonus. They can be used to motivate agents to explore new states within their influence.  Similarly to Equ. (\ref{Equ:GCR}), the exploration bonus to individual $\pi_i^e$ is defined as:
\begin{equation} \label{Equ:ER}
	R_{i+}(s, a_i, s') =
	\begin{cases}
		R_{i+}^c(s') + \beta_1  R_{i+}^{(i-c)}(s') &\text{if}  \quad{D(a_i) \cap D^c \neq \emptyset}\\
		\beta_1 R_{i+}^{(i-c)}(s') &\text{Otherwise} \\
	\end{cases}
\end{equation}
where $\beta_1$ is the scaling factor. Since $R_{i+}^c(s')$ and $R_{i+}^{(i-c)}(s')$ in Equ. \ref{Equ:ER} only involve counting over $s'^c$ and $s'^{(i-c)}$ delimited by $D_i$, it encourages agents to focus on discovering novel states within their respective influence scopes, thereby achieving more efficient exploration.
% Finally, exploration policies $\{\pi^e_i\}_{i\in\mathcal{I}}$ are trained with the combinations of exploration and environmental rewards: $r_i + \beta_2 r$, where $r^c_{i+} = R_{i+}(s, a_i, s')$, $r$ is the environmental reward and $\beta_2$ is a scaling factor.

% $r^+(s) = \beta / \sqrt{N(\varphi(s))}$

% \footnotesize 

We organize the pseudo-code of ISA in Algorithm \ref{alg1}. After the initialization in Line 1, the influence scope will be computed in Lines 2$\sim$14 before training. During training, $\{\pi^e_i\}_{i\in\mathcal{I}}$ is first trained to discover successful states in Lines 19$\sim$24. When enough goals are collected (i.e., $len(\mathcal{B}) \ge L$ where $len(\mathcal{B})$ represents the length of buffer $\mathcal{B}$. We implement $\mathcal{B}$ as a set), goal-conditioned policies $ \{\pi_i\}_{i\in\mathcal{I}}$ will learns to solve the task in the given Dec-POMDP in Lines 26$\sim$32. During interaction, success states found by $\{\pi^e_i\}_{i\in\mathcal{I}}$ are stored in goal buffer $\mathcal{B}$ in Line 34$\sim$36. We use the IPPO loss \citep{IPPO} to train $\{\pi^e_i\}_{i\in\mathcal{I}}$ and $\{\pi_i\}_{i\in\mathcal{I}}$ separately with their corresponding rewards.

For updating $\{\pi^e_i\}_{i\in\mathcal{I}}$, we strictly follow the IPPO loss \citep{IPPO}. For updating $\{\pi_i\}_{i\in\mathcal{I}}$, we extended IPPO by augmenting the observation of policies and value functions with individual goals. Specifically, the loss of value function is
\begin{equation}\label{critic_i}
	\mathcal{L}_{i}(\phi) = \mathbb{E}_{\tau_i^{t}}\biggl[ \min \biggl\{\biggl( V^{\phi}_i (\tau_i^{t}, g_i) - \hat{V}^{\phi}_i \biggr)^2, \biggl( V^{\phi^{old}}_i (\tau_i^{t}, g_i) + V^{clip}_i - \hat{V}^{\phi}_i  \biggr)^2   \biggr\} \biggl]
\end{equation}where $\tau_i^{t} \in \mathcal{T}_i$ is the current trajectory of agent $i$, $V^{\phi}_i$, $V^{\phi^{old}}_i$ and $V^{clip}_i$ are the value functions in IPPO \citep{IPPO}. Here, the only difference from IPPO is that our value function takes not only the trajectory $\tau_i^{t}$ as input but also the individual goal $g_i$, which is used to estimate the value of the individual trajectory for the individual goal. The policy loss for agent $i$ is
\begin{equation}
	\mathcal{L}_{i}(\theta) = \mathbb{E}_{\tau_i^{t}, a_i^{t}}\biggl[ \min \biggl\{ \frac{\pi_i^{\theta}(a_i^{t}|\tau_i^{t}, g_i)}{\pi_i^{\theta^{old}}(a_i^{t}|\tau_i^{t}, g_i)} A^t_{i}, clip(\frac{\pi_i^{\theta}(a_i^{t}|\tau_i^{t}, g_i)}{\pi_i^{\theta^{old}}(a_i^{t}|\tau_i^{t}, g_i)}, 1-c,1+c) A^t_{i} \biggr\} \biggl]
\end{equation}
where $\pi_i^{\theta}$ represents the $\pi_i$ parameterized by $\theta$, $\pi_i^{\theta^{old}}$ is the old policy before the update, $A^t_{i}$ is the IPPO advantage, $clip$ is a function to keep the ratio within the trust region $(-c, +c)$ and $c$ is hyperparameter \citep{IPPO}. The overall learning loss is
\begin{equation} 
	\mathcal{L} = \sum_{i=1}^{|\mathcal{I}|} \biggl(\mathcal{L}_{i}(\theta) + \lambda_{critic} \mathcal{L}_{i}(\phi) + \lambda_{entropy} \mathcal{H}(\pi_i^{\theta}) \biggr)
\end{equation}
where $\mathcal{H}(\pi_i^{\theta})$ denotes the entropy of policy $\pi_i^{\theta}$.

\begin{algorithm}[H]  \footnotesize
	\caption{Influence Scope of Agents (ISA) .}
	\begin{algorithmic}[1]
		\State Initialize random exploration policies $\{\pi^e_i\}_{i\in\mathcal{I}}$, goal-conditioned policies $ \{\pi_i\}_{i\in\mathcal{I}}$ and  goal buffer $\mathcal{B}$
		\For{$i \in \mathcal{I}$}
		\State Begin a new episode and randomly select $\bm{a_{-i}}$
		\For{$n = 1$ to $N$}
		\State Collect transitions with random $a_i$ and fixed $\bm{a_{-i}}$
		\If{$Env$ terminal}
		\State Begin a new episode and randomly select $\bm{a_{-i}}$
		\EndIf
		\EndFor
		\For{$a_i\in A_i$}
		\State Estimate $I(\Delta s^k; a | \bm{a_{-i}})$ by Equation (\ref{Equ:I}) with $N$ collected transitions
		\State Calculate $D(a_i)$ with Definition 1
		\EndFor
		\EndFor
		\State Calculate $D_i$, $D^c$, $D^{(i-c)}$ according to Definition \ref{Def:inf_ag} and \ref{Def:segments}
		\For{Episode $1$ to $M$}
		\State Begin a new episode 
		\If{$len(\mathcal{B}) < L$}
		\State Collect a trajectory with $\{\pi^e_i\}_{i\in\mathcal{I}}$
		\For{$(s,\bm{a},s',r) \in$ trajectory}
		\State Count $\varphi(s'^{(i-c)})$ for each $i \in \mathcal{I}$ and $\varphi(s'^c)$
		\State Obtain rewards $R_{i+}(s, a_i, s')$ for each $i \in \mathcal{I}$ with Equation (\ref{Equ:ER})
		\State Update $\pi^e_i, \forall i \in \mathcal{I}$ with IPPO loss and $R_{i+}(s, a_i, s') + \beta_2 r$
		\EndFor
		\Else
		\State Sample a state as global goal $g$ from $\mathcal{B}$
		\State Decompose $g$ into $\{g_i\}_{i\in \mathcal{I}}$ based on Definition 3
		\State Collect a trajectory with $\{\pi_i\}_{i\in\mathcal{I}}$ and $\{g_i\}_{i\in \mathcal{I}}$
		\For{$(s,\bm{a},s',r) \in$ trajectory}
		\State Calculate rewards $R_i(s, a_i, s' | g_i)$ for each $i \in \mathcal{I}$ with Equation (\ref{Equ:GCR})
		\State Update $\pi_i, \forall i \in \mathcal{I}$ with IPPO loss and $R_i(s, a_i, s' | g_i) + \alpha_2 r$
		\EndFor
		\EndIf
		\If{trajectory is success}
		\State Store terminated state into $\mathcal{B}$
		\EndIf
		\EndFor
	\end{algorithmic}
	\label{alg1}
\end{algorithm}

\section{Experiments}
\label{Sec:exp}

\textbf{Environment.} We evaluate our method in two challenging sparse-reward benchmarks: (1) the starcraft multi-agent challenge (SMAC) \citep{SMAC}; and (2) the multiple-particle environment (MPE) \citep{MPE}. 

\begin{wraptable}{r}{0.5\linewidth} 
    \centering
    \footnotesize
    \renewcommand{\arraystretch}{1.05}

    \begin{tabular}{c|c|c|c}
        \hline
        Event & 
        \begin{tabular}[c]{@{}c@{}}Dense \\ Reward\end{tabular} &
        \begin{tabular}[c]{@{}c@{}}Sparse \\ Reward\end{tabular} &
        \begin{tabular}[c]{@{}c@{}}Super Sparse \\ Reward\end{tabular} \\
        \hline
        Enemy hit-point $h_e$ & -$h_e$ & 0 & 0 \\
        Ally hit-point $h_a$  & +$h_a$ & 0 & 0 \\
        An enemy death        & +10 & +10 & 0 \\
        An ally death         & -5  & -5  & 0 \\
        Win                   & +200 & +200 & +1 \\
        Loss                  & 0 & 0 & -1 \\
        \hline
    \end{tabular}

    \vspace{2mm}
    \caption{The reward setting for SMAC domain.}
    \label{tab:smac_reward}
\end{wraptable}

In SMAC, agents need to learn cooperative policies to eliminate the enemies. SMAC has different reward setting as shown in Table \ref{tab:smac_reward}. We focus on the super sparse setting where a `+1' reward will be given only when all enemies are eliminated and a `-1' reward will be given when all controlled agents die or episode terminates with alive enemies. This setting is known as particularly difficult due to the absence of intermediate rewards \citep{HMASD}. Besides, in line with the pervious work \citep{HMASD, QMIX, FoX}, we set the difficulty level to 7. Following \citep{HMASD}, we use SMAC instead of SMACv2 \citep{smacv2} to evaluate the algorithms in sparse-reward scenarios.

In addition to SMAC, we also consider 3 tasks in MPE, i.e., Navigation, Shooting and Unlock. In the Navigation task, agents need to learn to occupy different landmarks where a `+1' reward is given to the team when a landmark is occupied. In the Shooting task, agents need to shoot a target enough times and move towards specific positions where a `+1' reward is given only when these subtasks finished. In the Unlock task, agents holds different keys to unlock the corresponding (multiple) locks where a `+1' reward is given to the team when a lock is unlocked. 

\textbf{Baselines.} Our baselines cover classical MADRL methods (i.e., IPPO \citep{IPPO}, QMIX \citep{QMIX}, COMA \citep{COMA}), methods with similar distance-based intrinsic rewards to ISA (i.e., MASER \citep{MASER}), and state-of-the-art methods in sparse-reward MADRL domains (i.e., CMAE \citep{CMAE}, FoX \citep{FoX}, HMASD \citep{HMASD}). More details about environments and baselines are included in Appendix B.1$\sim$B.2.

\textbf{Hyperparameters.} We run ISA on 2.60 GHz AMD Rome 7H12 CPU. The hyperparameter settings for the learning part of ISA make reference to IPPO \citep{IPPO}. For the introduced hyperparameter $\delta$, we fine-tune its value to find the workable range. We observed when $\delta \in [0.15,0.45]$, the identification is correct for almost all actions across domains. We set $\delta=0.3$ in all tasks and this works well. The scaling factors $\alpha_1$ and $\beta_1$ in Equ. (\ref{Equ:GCR}) and (\ref{Equ:ER}) are selected from $\{0, 0.2\}$ across domains. See Appendix B.3 for the analysis for the empirical choice of these hyperparameters. The number of transitions to calculate influence scopes is 10,000 in 8m and 2,000 in other environments. The length $L$ in Algorithm \ref{alg1} is 1 for all environments.

\begin{figure*}[!t]
    \centering

    % ---- Row 1 (4 plots) ----
    \begin{subfigure}[t]{0.21\textwidth}
        \centering
        \includegraphics[width=\linewidth]{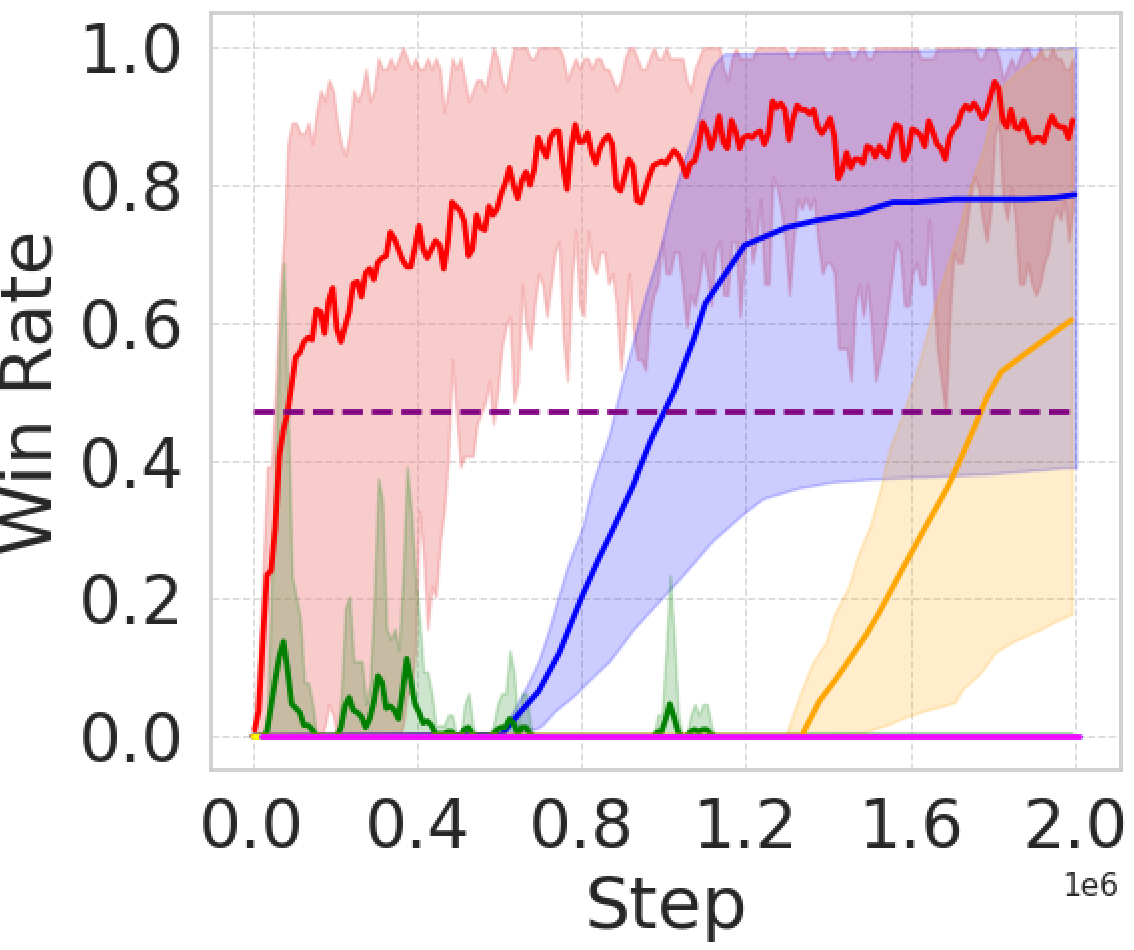}
        \caption{3m}
    \end{subfigure}\hfill
    \begin{subfigure}[t]{0.21\textwidth}
        \centering
        \includegraphics[width=\linewidth]{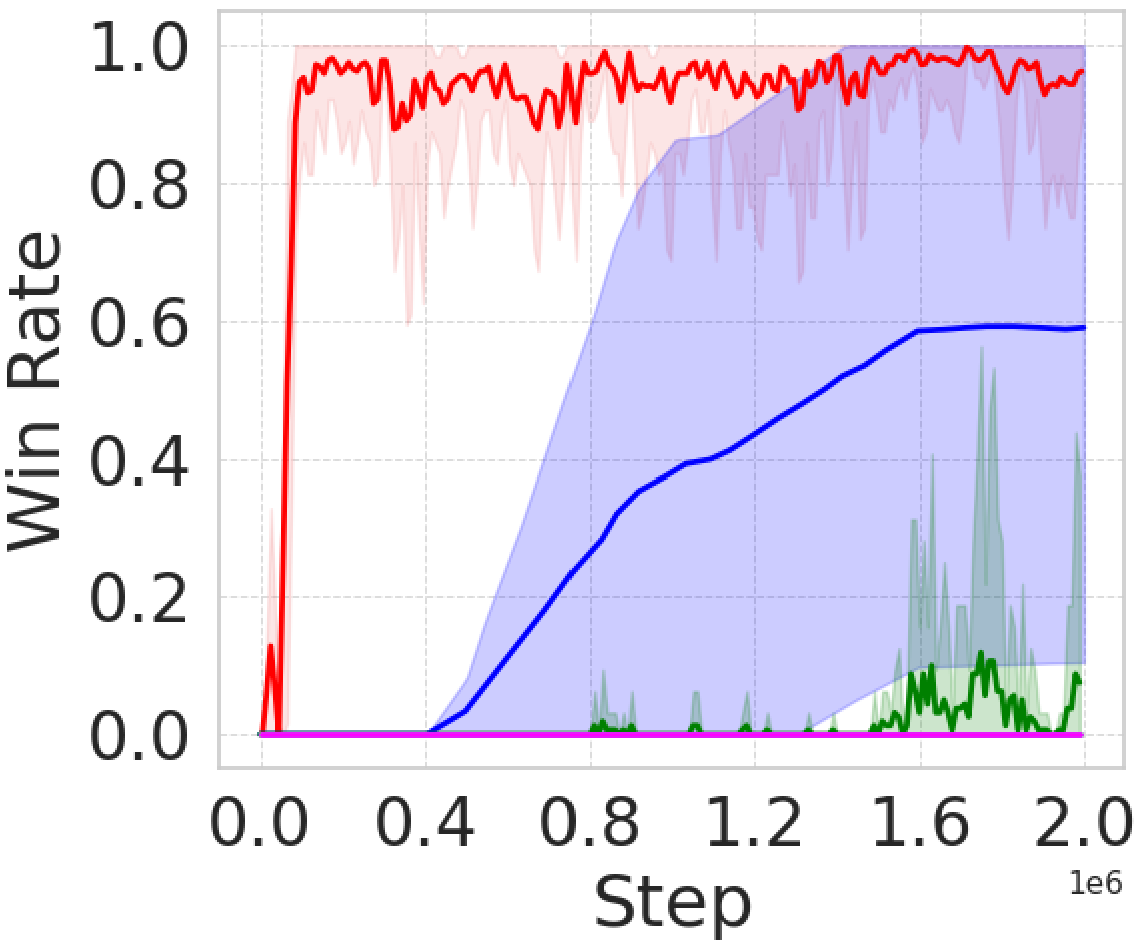}
        \caption{2s\_vs\_1sc}
    \end{subfigure}\hfill
    \begin{subfigure}[t]{0.21\textwidth}
        \centering
        \includegraphics[width=\linewidth]{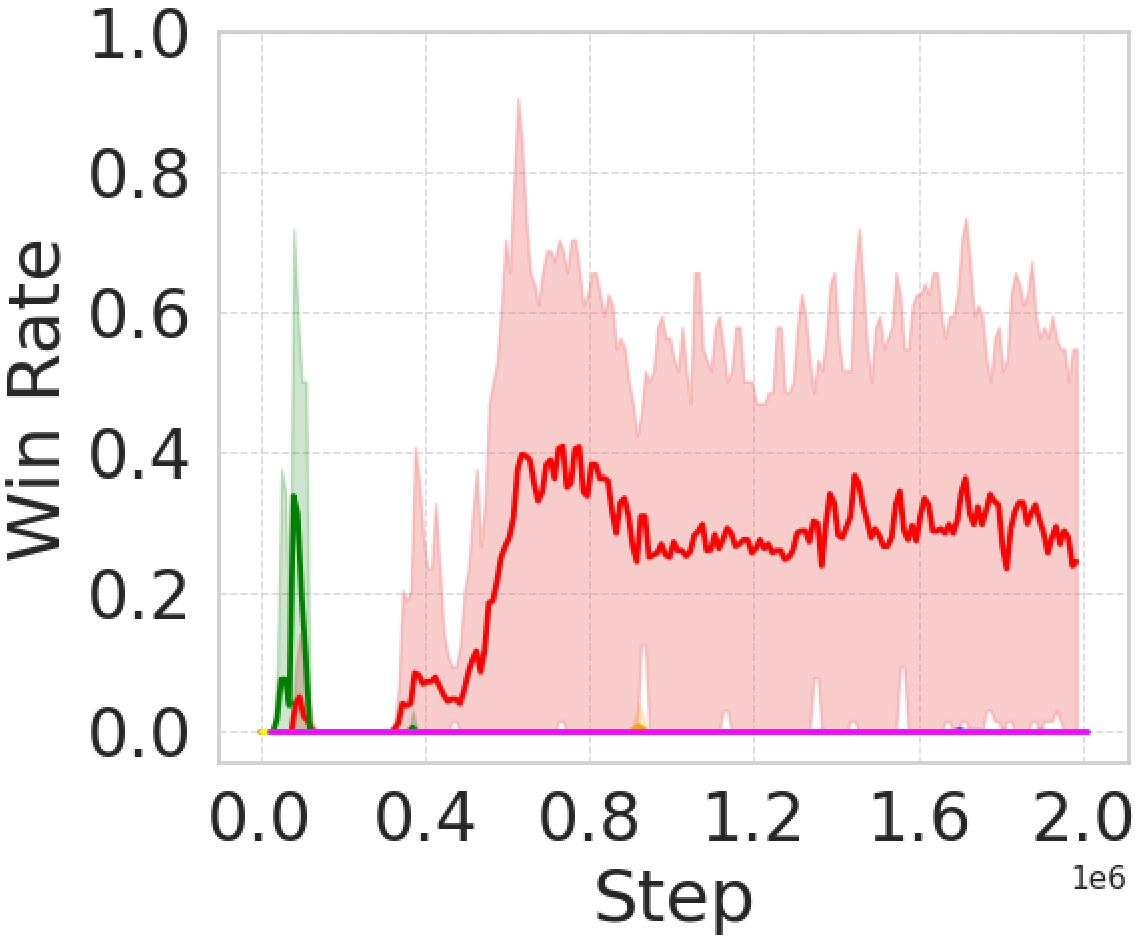}
        \caption{8m}
    \end{subfigure}\hfill
    \begin{subfigure}[t]{0.21\textwidth}
        \centering
        \includegraphics[width=\linewidth]{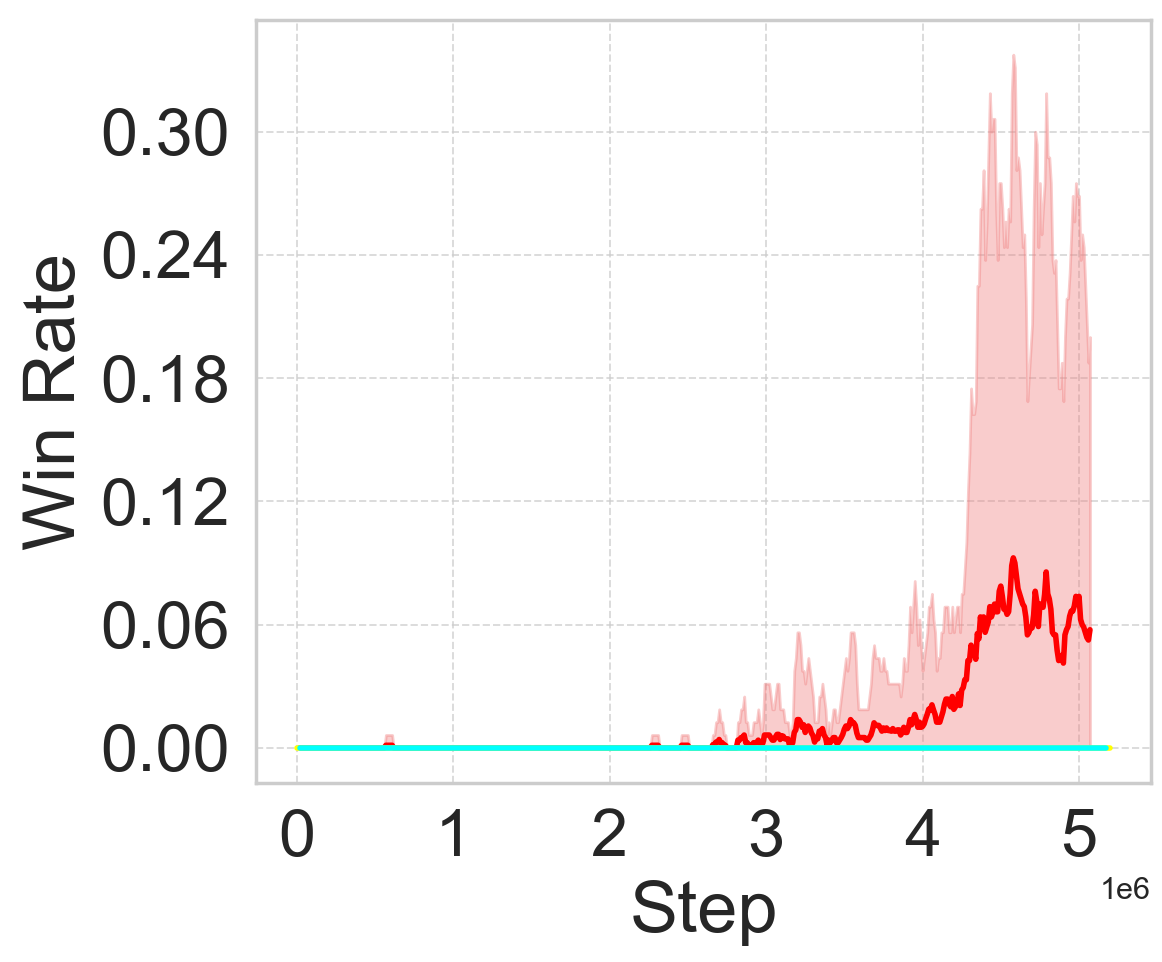}
        \caption{8m\_vs\_9m}
    \end{subfigure}

    % ---- Row 2 (4 plots) ----
    \begin{subfigure}[t]{0.21\textwidth}
        \centering
        \includegraphics[width=\linewidth]{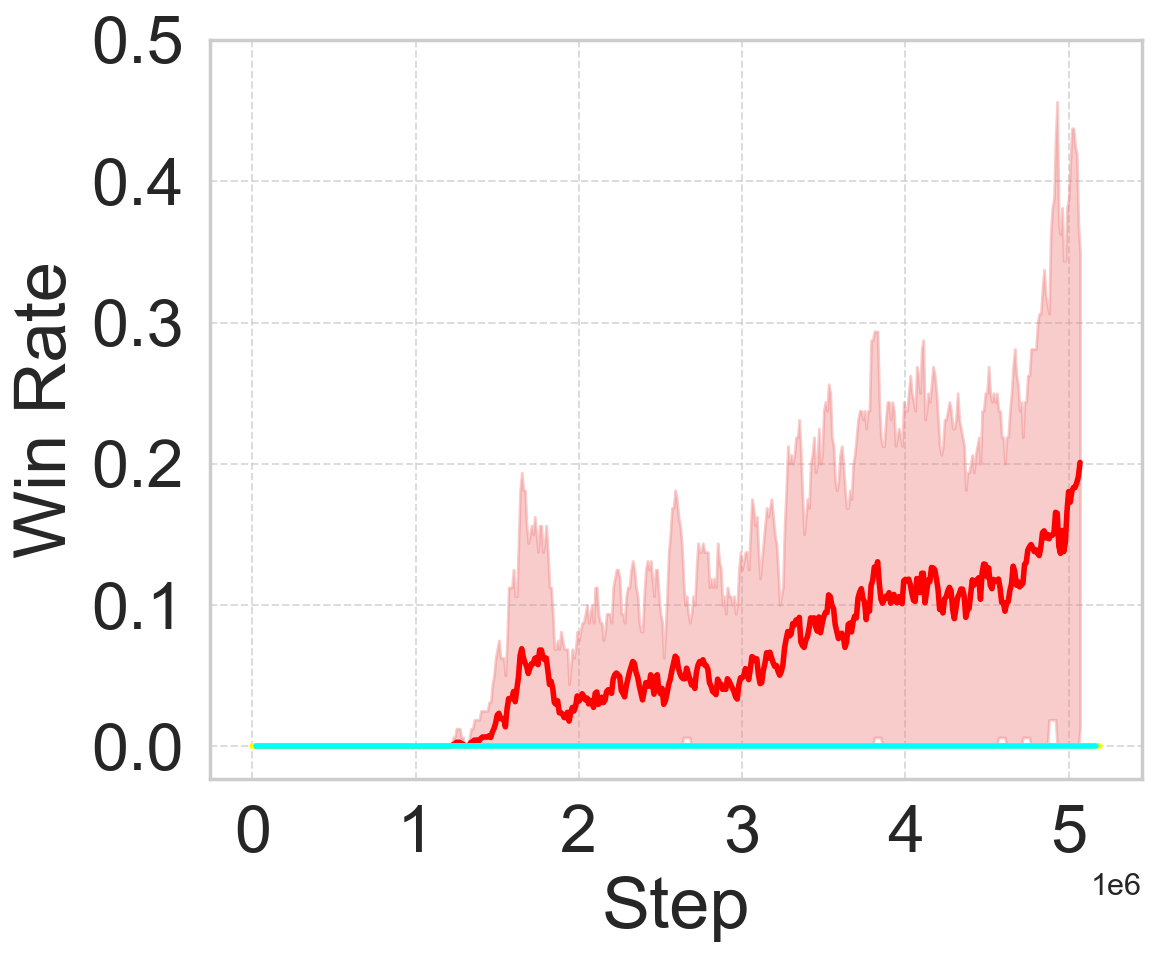}
        \caption{10m\_vs\_11m}
    \end{subfigure}\hfill
    \begin{subfigure}[t]{0.21\textwidth}
        \centering
        \includegraphics[width=\linewidth]{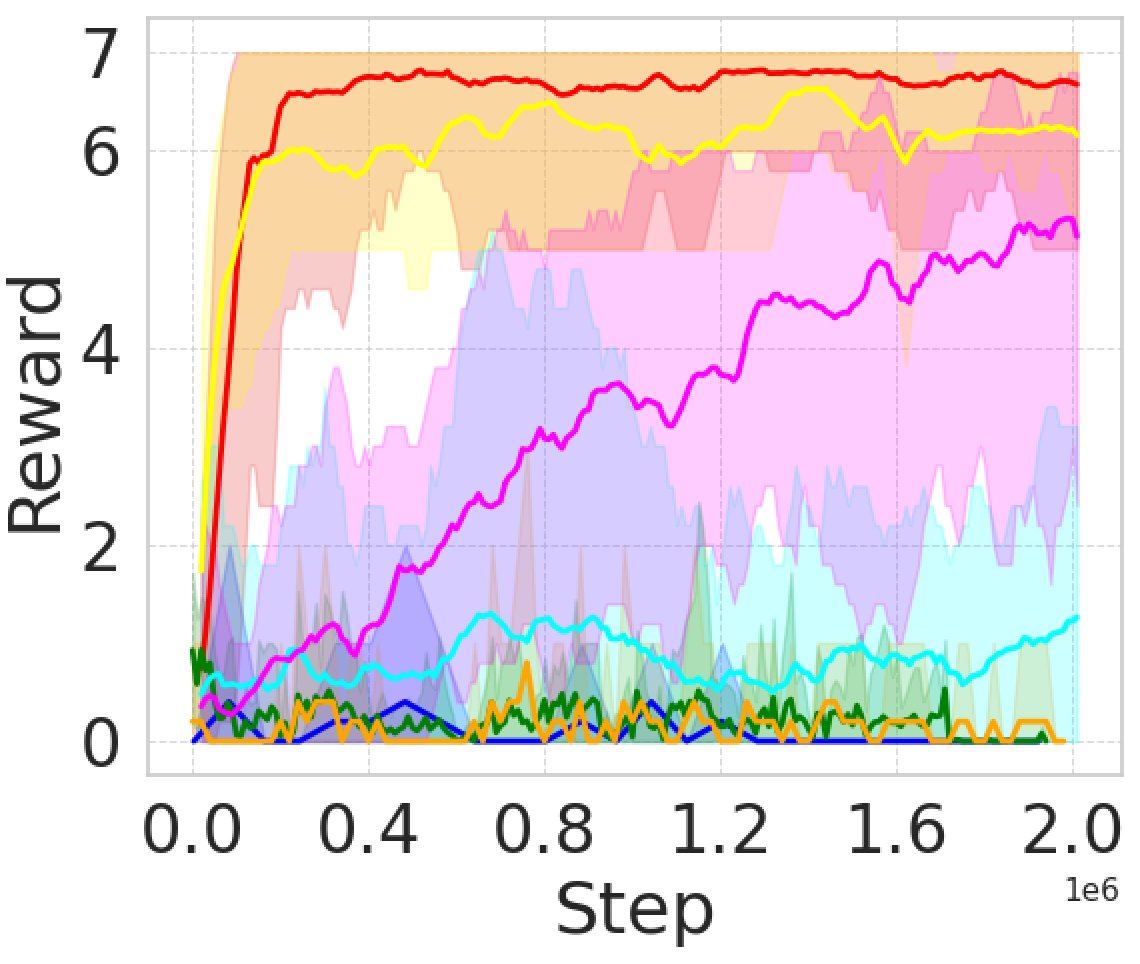}
        \caption{Navigation}
    \end{subfigure}\hfill
    \begin{subfigure}[t]{0.21\textwidth}
        \centering
        \includegraphics[width=\linewidth]{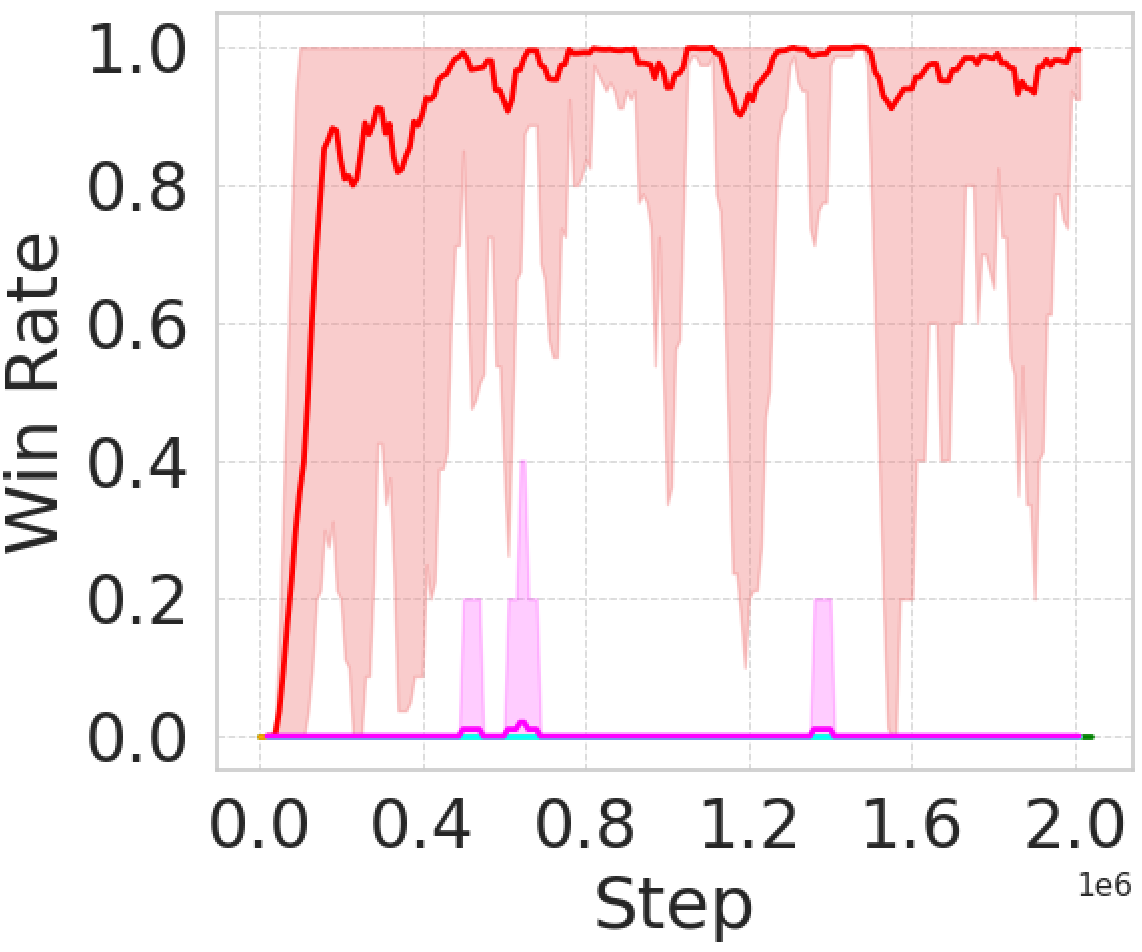}
        \caption{Shooting}
    \end{subfigure}\hfill
    \begin{subfigure}[t]{0.21\textwidth}
        \centering
        \includegraphics[width=\linewidth]{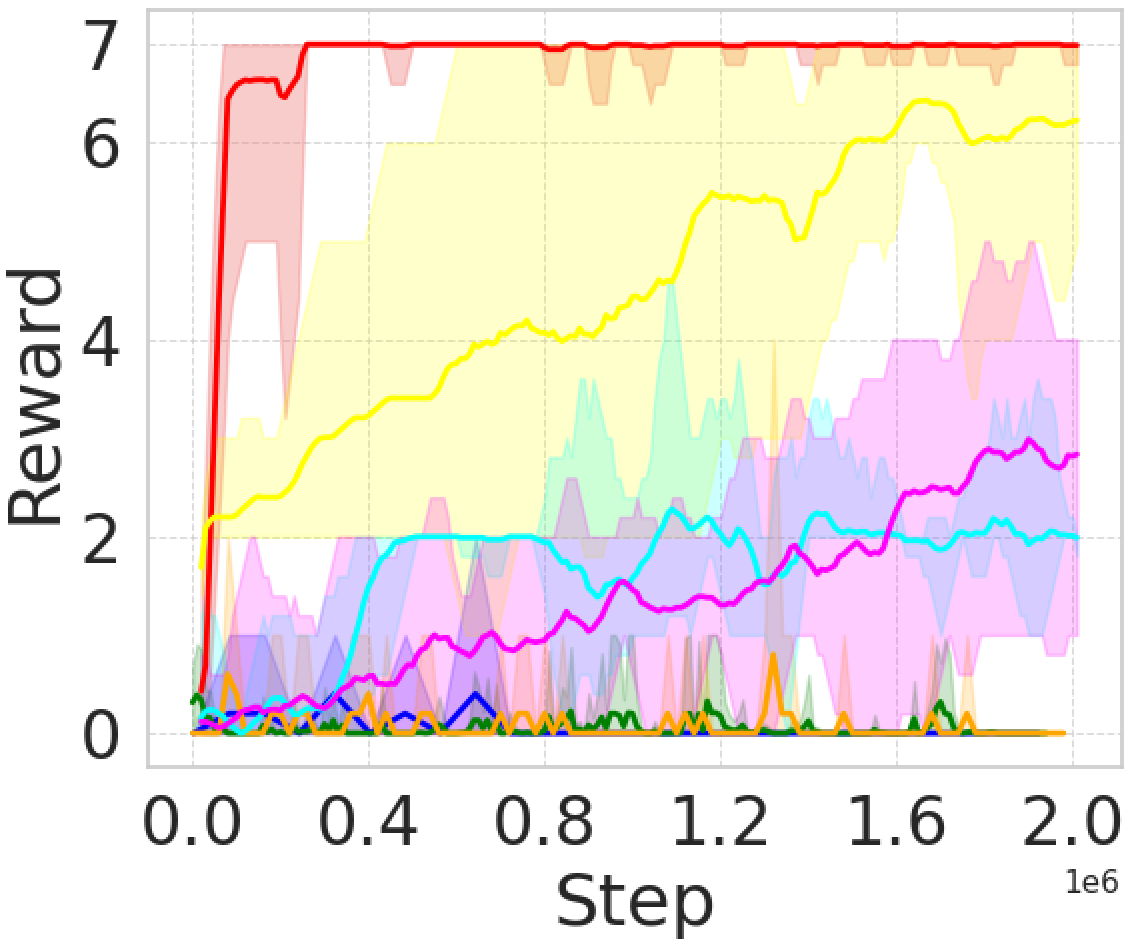}
        \caption{Unlock}
    \end{subfigure}

    % ---- Legend (full width) ----
    \begin{subfigure}[t]{0.92\textwidth}
        \centering
        \includegraphics[width=\linewidth]{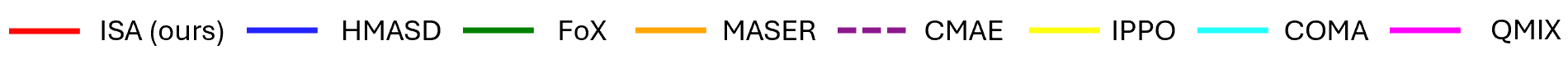}
        \caption*{} % 不编号不显示 caption（需要就删掉这行）
    \end{subfigure}

    \vspace{-2mm}
    \caption{Learning curves on SMAC (with only a $+1/-1$ reward at terminate state) and MPE.}
    \label{fig:compare}
\end{figure*}

\begin{figure*}[!h]
	\centering
	\includegraphics[width=14cm]{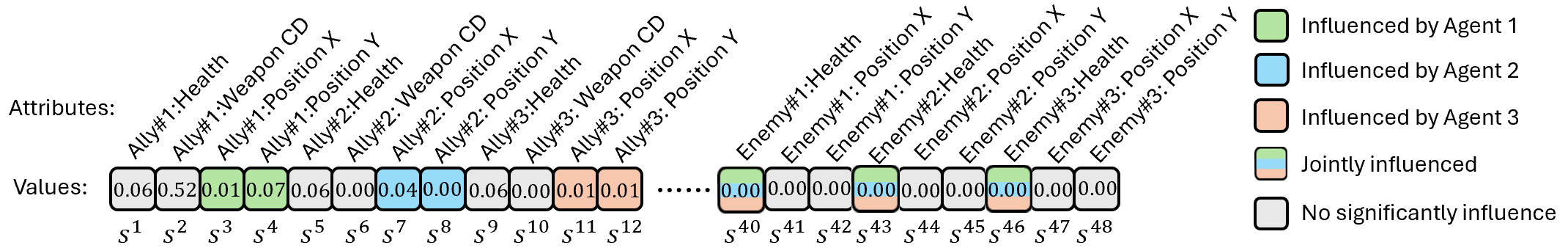}
	\caption{An illustration on decomposing individual goals from a global goal.}
	\vspace{-3mm}
	\label{fig:goald}
\end{figure*}

\textbf{Results.} We compare our method with baselines on 8 environments from SMAC and MPE domains to validate the superiority of ISA. The results are shown in Fig. \ref{fig:compare}. The error bounds (i.e., shadow shapes) indicate the upper and lower bounds of the performance with 5 runs using different random seeds. For a fair comparison, all environment interactions of ISA are counted in the total training steps shown in Fig. \ref{fig:compare}. This includes the Monte Carlo sampling used for influence scope estimation in Step 1 as well as the exploration phase in Step 2. Due to the delayed effect of actions in 2s\_vs\_1sc, we take into account the state changes in the next 2 steps of the current action when calculating $D(a_i)$. As shown in Fig. \ref{fig:compare}, classical MADRL methods fails to learn performance with sparse rewards. FoX and MASER perform well in their original papers under partially sparse environments, but struggle to learn effectively in our extremely sparse reward settings. ISA consistently outperforms the state-of-the-art method HMASD under same settings. In particular, on 2s\_vs\_1sc, ISA achieves a 56\% improvement over the strongest baseline after 1.6M environment steps.

\textbf{Goal decomposition.} We visualize ISA’s goal decomposition in Fig. \ref{fig:goald}, where each square represents one of the 48 dimensions of a success state $s=[s^1,s^2,...,s^{48}]$ explored from the \textit{3m} task. Taking this state as a global goal, individual goals of agents are decomposed based on the dimensions/attributes that they can can influence. For instance, agent 1's goal consists of  the green (special) and colorful (common) segment, covering its own position and enemy health.

\begin{wrapfigure}{r}{0.6\linewidth}
    \vspace{-6mm}
    \centering

    \begin{subfigure}[t]{0.44\linewidth}
        \centering
        \includegraphics[width=\linewidth]{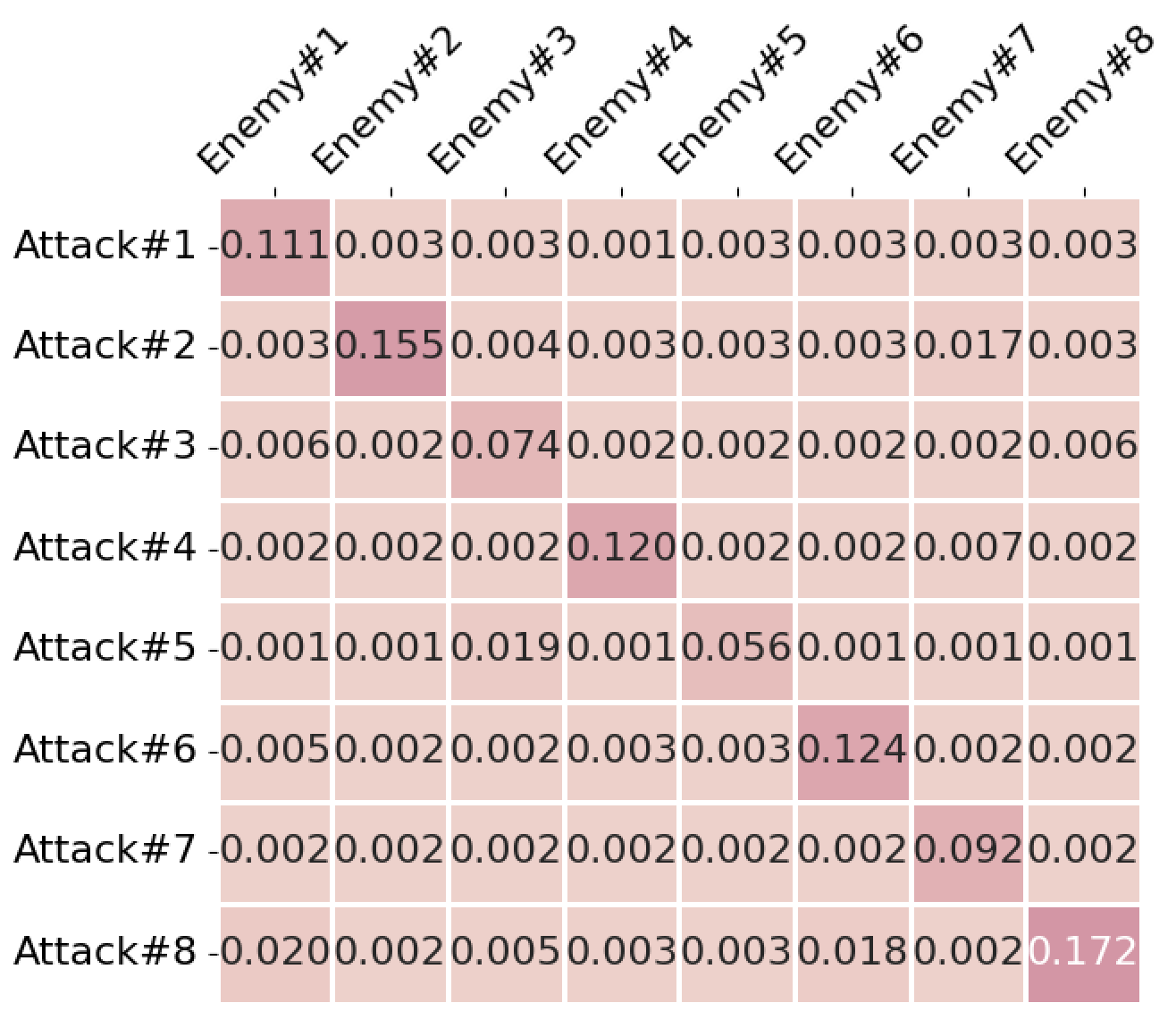}
        \caption{}
        \label{fig:ab_a}
    \end{subfigure}\hfill
    \begin{subfigure}[t]{0.5\linewidth}
        \centering
        \includegraphics[width=\linewidth]{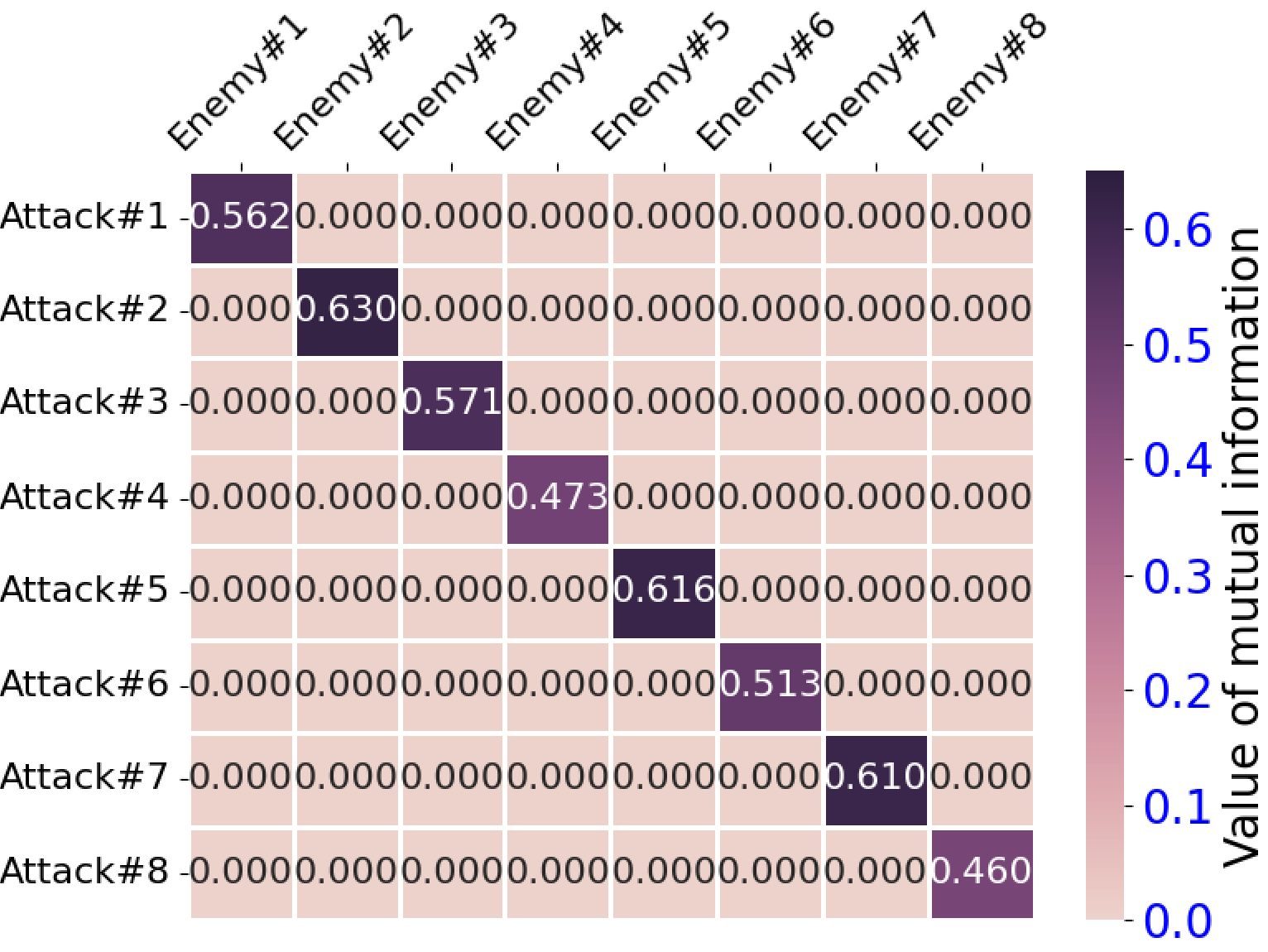}
        \caption{}
        \label{fig:ab_b}
    \end{subfigure}

    \vspace{2mm}

    \begin{subfigure}[t]{0.45\linewidth}
        \centering
        \includegraphics[width=\linewidth]{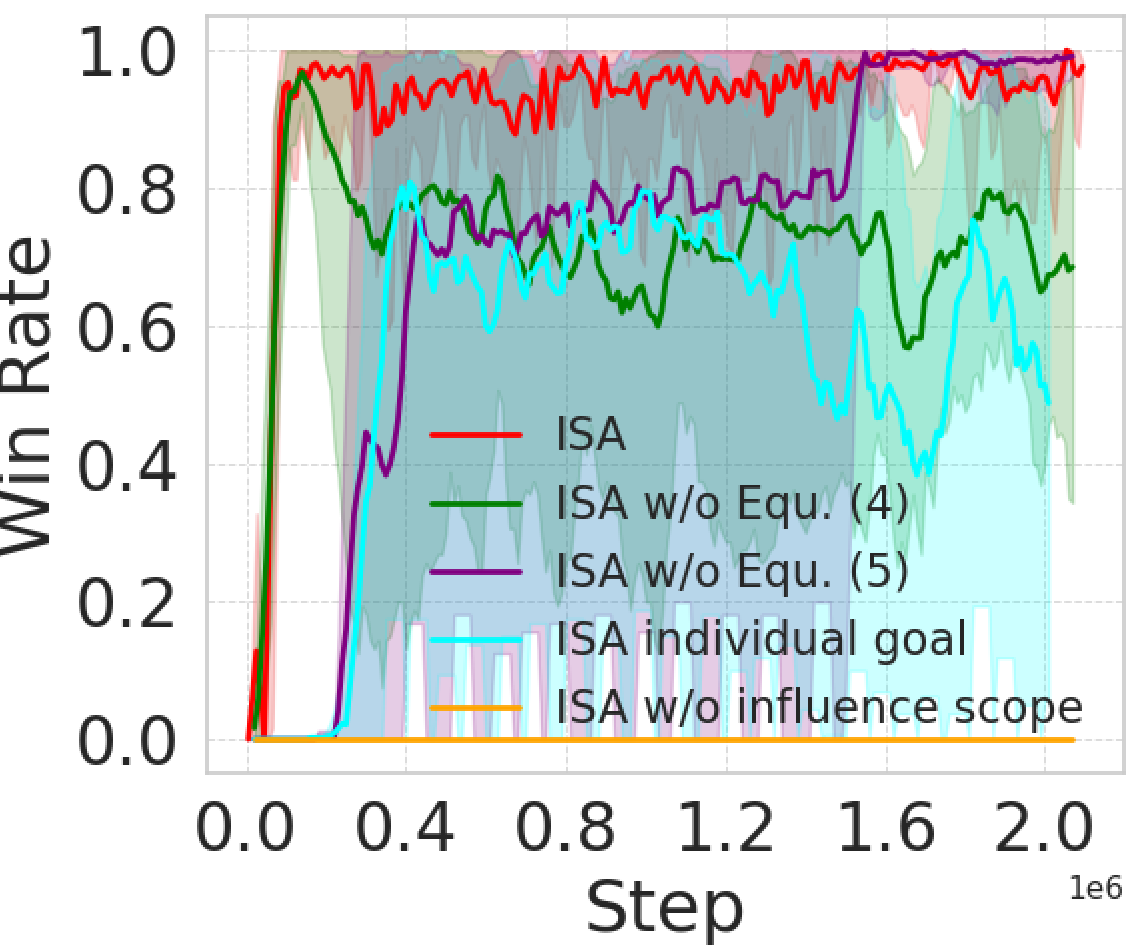}
        \caption{}
        \label{fig:ab_p}
    \end{subfigure}\hfill
    \begin{subfigure}[t]{0.44\linewidth}
        \centering
        \includegraphics[width=\linewidth]{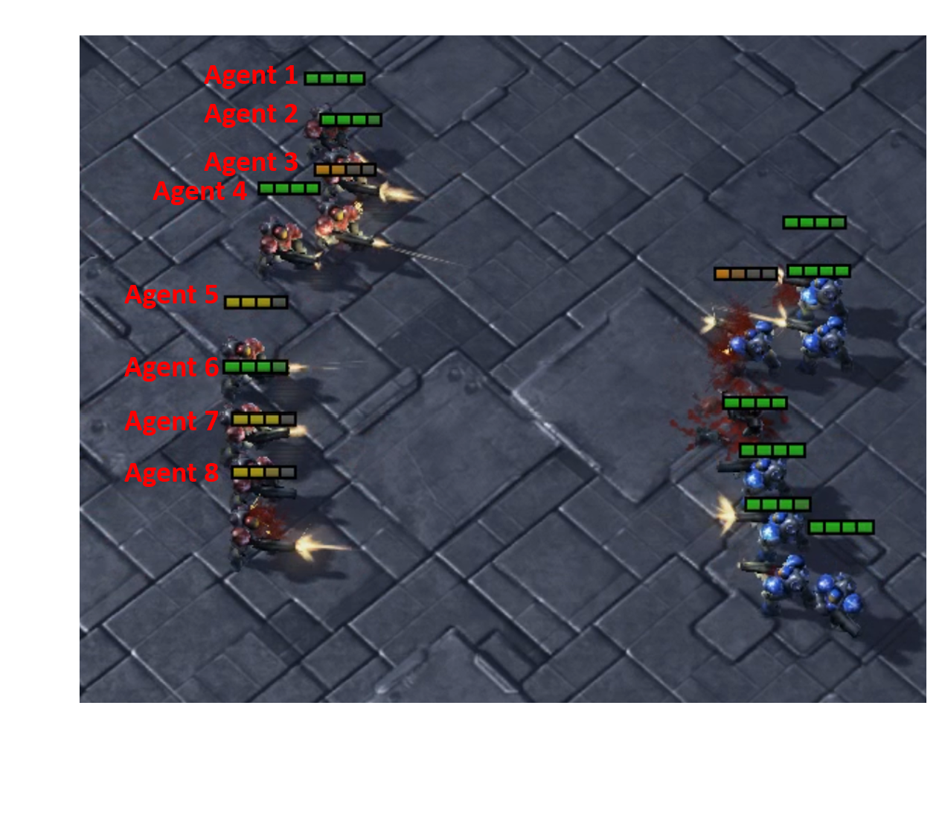}
        \caption{}
        \label{fig:ana_d}
    \end{subfigure}

    \vspace{-2mm}
    \caption{
    Heat maps of mutual information between action and state change 
    (a) not conditioned on $\bm{a_{-i}}$ and 
    (b) conditioned on $\bm{a_{-i}}$. 
    (c) Learning curves of different versions of ISA in 2s\_vs\_1sc. 
    (d) Screenshot in 8m for interpretability.
    }
    \label{fig:ab}
    \vspace{-2mm}
\end{wrapfigure}

\textbf{Ablations.} We compare the mutual information in Equation (\ref{Equ:I}) with and without conditioning on $\bm{a_{-i}}$ in Equ. (\ref{Equ:I}), as shown in Fig. \ref{fig:ab}.
The values in Fig. \ref{fig:ab} represent the values of mutual information. For example, the the value on (3,3) position in Fig. \ref{fig:ab}a indicates mutual information between agent 1’s action ``attack enemy 3” and the change in enemy 3’s health, without conditioning on \( \bm{a_{-i}} \); Correspondingly, the value on the same position in Fig. \ref{fig:ab}b represents the same mutual information but conditioned on $\bm{a_{-i}}$. The results show that the influence on state of the agent's actions can be distinguished more clear when conditioned on $\bm{a_{-i}}$. 

Besides, we perform ablations to verify the contribution of influence scope and the effectiveness of credit assignment in Fig. \ref{fig:ab}c. `ISA w/o influence scope' ablates the influence scope in ISA (preserving the count-based exploration over all state dimensions), which fails to learn due to the large exploration space. `ISA individual goal' ablates the segmentation on individual goals (constructing the reward based on the distance between current state and individual goals), which shows lower sample efficiency and instability because of the wrong credit assignment during both exploration and learning. `ISA w/o Equ. (4)' ablates the judging process for credit assignment in Equation (\ref{Equ:GCR}) based on the influence of action, which shows instability of policy learning due to wrong reward. `ISA w/o Equ. (5)' ablates the judging process for credit assignment in Equation (\ref{Equ:ER}) in exploration, which shows lower sample efficiency to find the success states to start goal-conditioned policy learning.

\textbf{Interpretability.} Our credit assignment based on the influence scope offers good interpretability. Based on the if-else rule in Equ. (\ref{Equ:GCR}), we can interpret whether a specific action $a_i$ of agent $i$ has influence on the common segment $D_c$. For instance, when $D(a_i)\cap D^c=\emptyset$, the current action $a_i$ has no influence on $D_c$, and as a result, no reward from common segment shall be assigned to agent $i$. Fig. \ref{fig:ab}d illustrates this scenario. At the time step of this screenshot, all agents are shooting to enemies except agent 7. This indicates that the current action of agent 7 does not contribute to the state changes delimited by $D_c$ (the health of all enemy). Consequently, ISA ensures a fair credit assignment by awarding agent 7 less rewards than the other agents.

\begin{figure}[!h] 
	\centering 
    \begin{subfigure}[2s\_vs\_1sc]{0.25\textwidth}
        \centering
        \includegraphics[width=\linewidth]{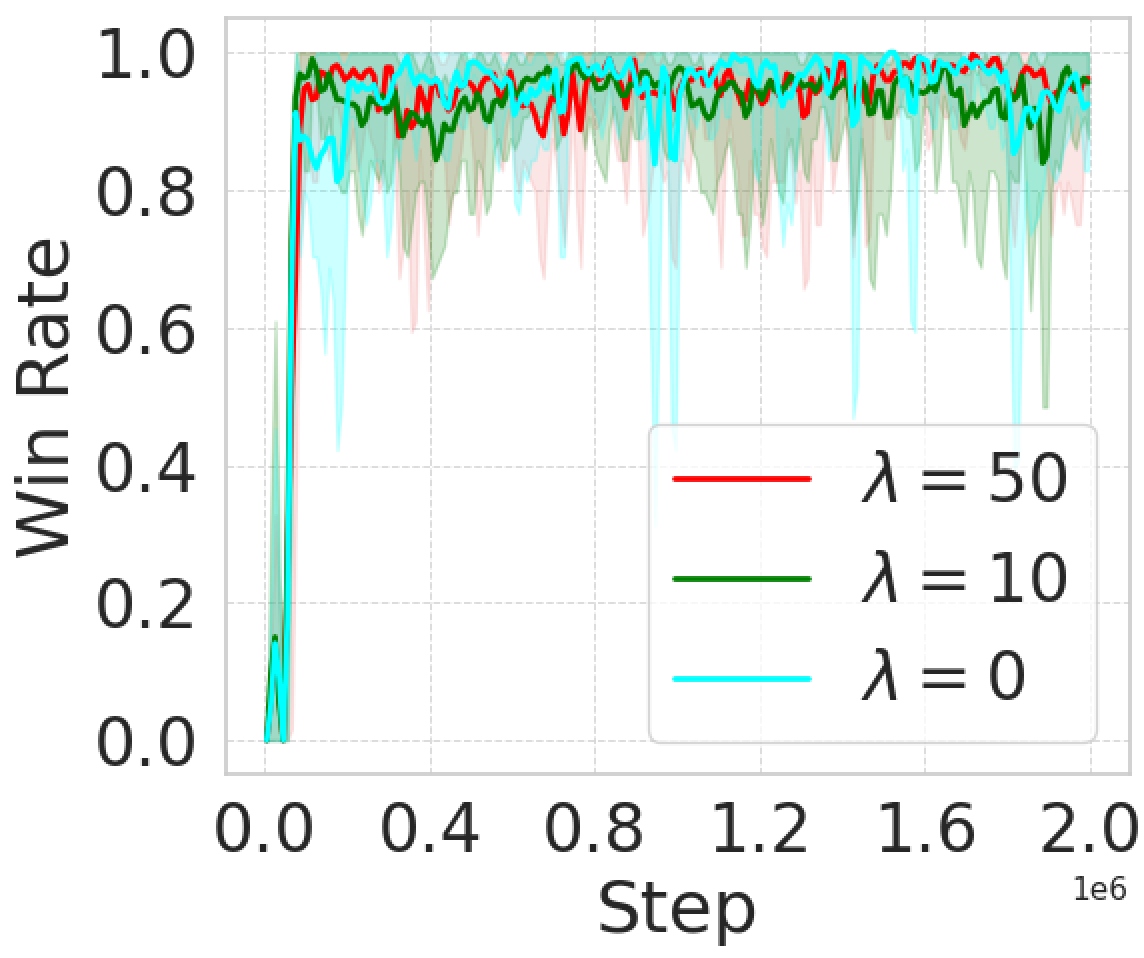}\label{fig:ab_a}
        \caption{2s\_vs\_1sc}
    \end{subfigure}
    \begin{subfigure}[3m]{0.25\textwidth}
        \centering
        \includegraphics[width=\linewidth]{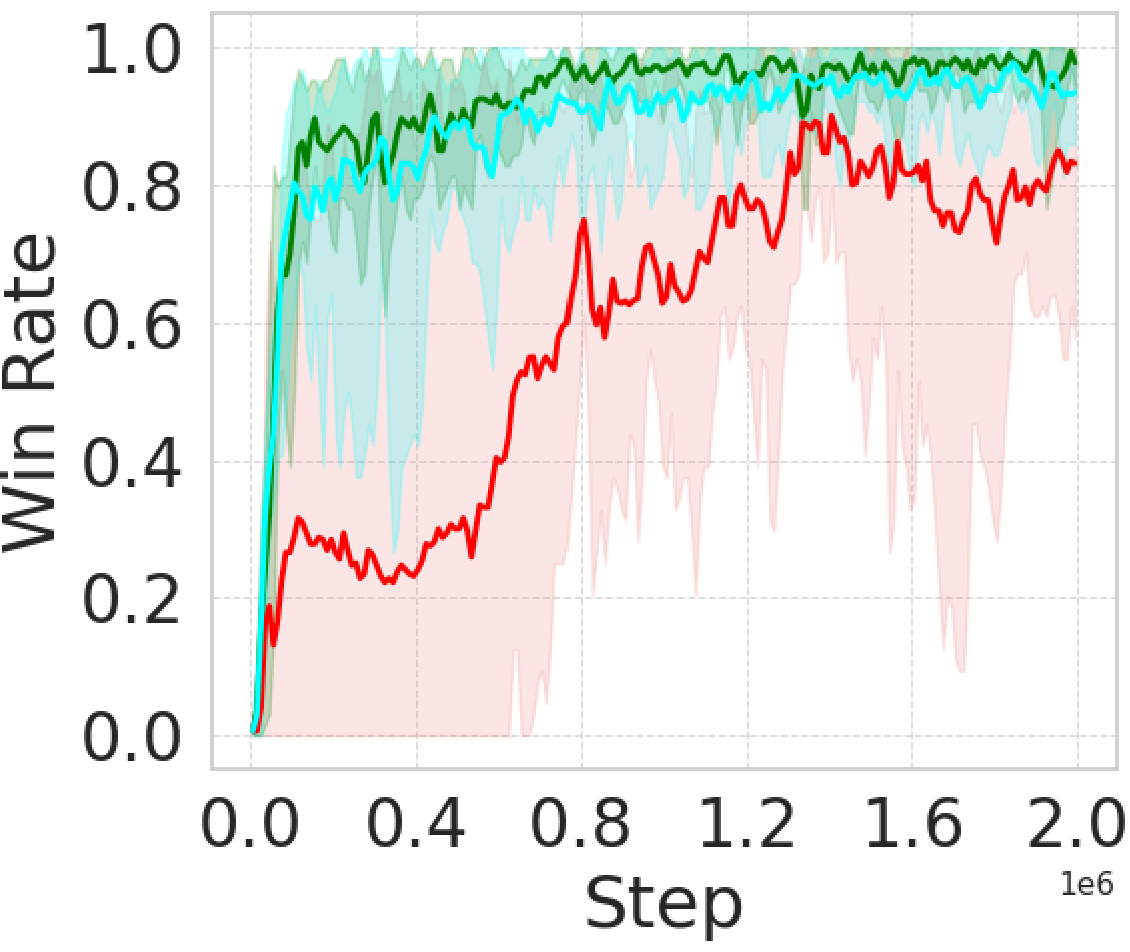}\label{fig:ab_b}
        \caption{3m}
    \end{subfigure}
    \begin{subfigure}[8m]{0.25\textwidth}
        \centering
        \includegraphics[width=\linewidth]{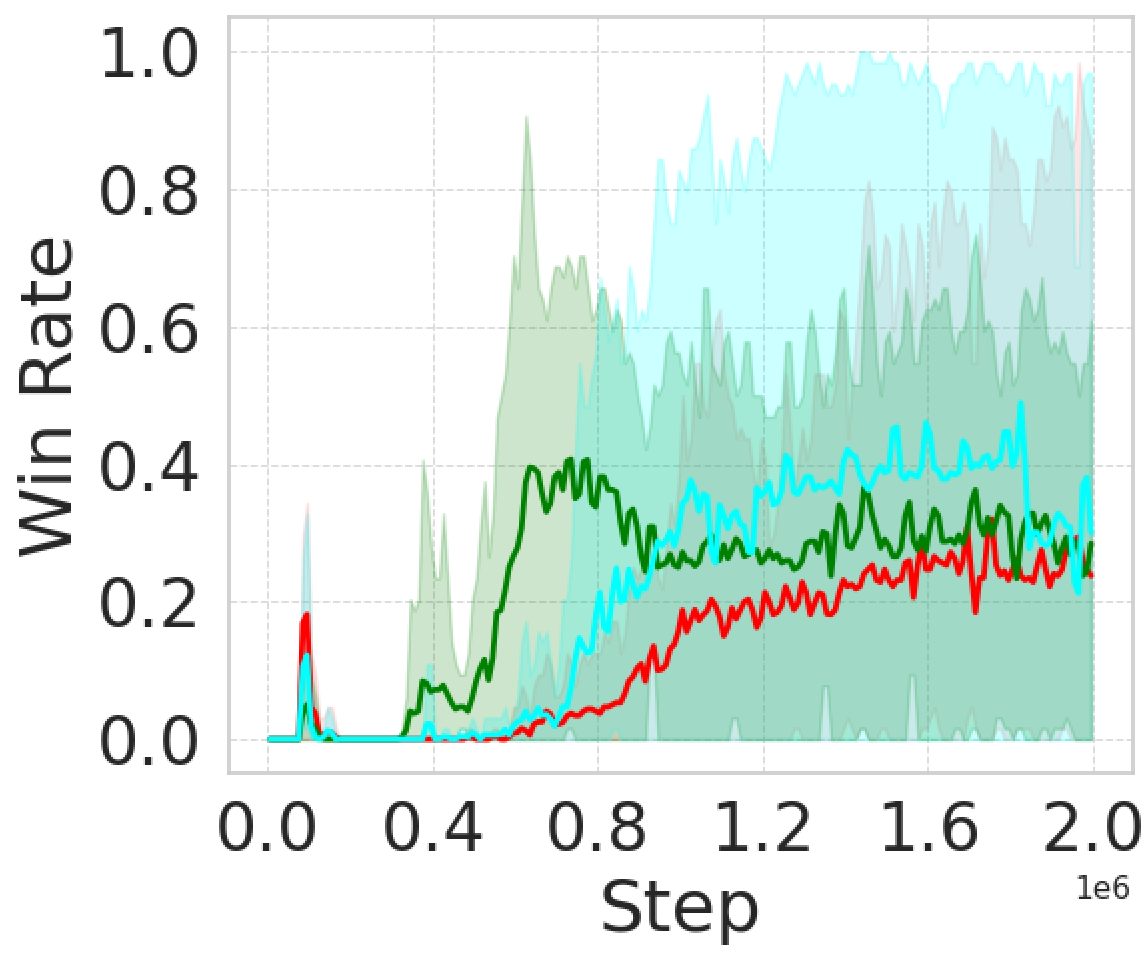}\label{fig:ab_c}
        \caption{3m}
    \end{subfigure}
	\caption{Performance of ISA with different $\lambda$.}
	\label{fig:hyper-hanming}
\end{figure}

\textbf{Hyperparameter Sensitivity.} We show the performance of ISA with different values of $\lambda$ in Figure \ref{fig:hyper-hanming}, where $\lambda$ is a key parameter in our method. It balances the Euclidean distance and the Hamming distance when computing the goal-conditioned reward. In this experiment, only the value of $\lambda$ is varied, while all other hyperparameters are fixed as listed in Table \ref{tab:hyper_b}. As shown in the figure, the performance of ISA remains relatively stable across a wide range of $\lambda$ values. While some performance fluctuations can be observed in specific environments, the overall trend indicates that ISA is robust to the choice of $\lambda$. This stability suggests that ISA does not rely heavily on precise tuning of this hyperparameter to achieve satisfactory performance, which enhances its practicality in diverse multi-agent settings.

\begin{wraptable}{r}{0.5\linewidth}
    \centering
    \footnotesize
    \renewcommand{\arraystretch}{1.05}

    \begin{tabular}{c|c|c|c}
        \hline
        & 3m & 2s\_vs\_1sc & 8m \\ 
        \hline
        Influence scope  & 3.71\%  & 3.67\%  & 3.74\%  \\
        Collect data     & 15.55\% & 15.49\% & 15.68\% \\
        Credit assignment & 2.70\% & 2.69\%  & 2.76\%  \\
        Training         & 78.04\% & 78.14\% & 77.82\% \\
        \hline
    \end{tabular}

    \vspace{2mm}
    \caption{Time consumption percentage of processes running $0.2$M steps for ISA in three scenarios.}
    \label{tab:time}
\end{wraptable}

\textbf{Introduced computation overhead.} In the experiments, we also observed that ISA has a low time overhead in the current challenging tasks. We record the proportion of computation time consumed by the following four processes during training. \textit{Influence scope} refers to the computation of influence scopes, which mainly corresponds to lines 2–15 in Algorithm 5, i.e., the calculation of $D^i$, $D^c$, and $D^{(i-c)}$. This process is introduced as a novel contribution of our work.
\textit{Collect data} denotes the interaction between agents and the environment during MADRL.
\textit{Credit assignment} refers to the computation of rewards based on Equation (5.4) and (5.5), which is also specific to our method.
\textit{Training} corresponds to the update of agents using IPPO.

Table \ref{tab:time} shows the time consumption percentage of ISA’s different processes in different scenarios. Compared to other goal-conditioned policy learning methods \citep{SurveyIntrinsically}, the possible computational overhead introduced by ISA is mainly involved in the calculation of the \textit{influence scope} in Definition 1$\sim$4 and the \textit{credit assignment} in Equation (\ref{Equ:GCR}) and (\ref{Equ:ER}) processes. As shown in Table \ref{tab:time}, the computation overhead percentage of these two processes is no more than 7 \% in these scenarios.

\section{Conclusions and future work}

In this paper, we propose ISA, an algorithm that improves both credit assignment and exploration in MADRL. ISA measures the mutual information between agents' actions and the state attributes/dimensions to identify the influence scope of agents. ISA use the influence scope to provide a precise and succinct representation for individual goals. Then, the credit assignment for the individual agent is determined based on the influence of its current action on its individual goal. Besides, a novel exploration method is proposed by restricting the state to be explored by agents to the attributes/dimensions of what they can influence, which improves the exploration efficiency.
We show that in a variety of sparse-reward MADRL environments, ISA significantly outperforms the state-of-the-art methods. In future work, we plan to explore the use of hierarchical influence-based goals in MADRL to improve policy learning.

%%%%%%%%%%%%%%%%%%%%%%%%%%%%%%%%%%%%%%%%%%%%%%%%%%%%%%%%%%%%%%%%
%% Bibliography
%%%%%%%%%%%%%%%%%%%%%%%%%%%%%%%%%%%%%%%%%%%%%%%%%%%%%%%%%%%%%%%%
\bibliography{main}
\bibliographystyle{rlj}

%%%%%%%%%%%%%%%%%%%%%%%%%%%%%%%%%%%%%%%%%%%%%%%%%%%%%%%%%%%%%%%%
% AUTHOR: If your paper has no supplementary materials, you may 
%         comment out the line below, which creates the title for
%         the supplementary materials.
%%%%%%%%%%%%%%%%%%%%%%%%%%%%%%%%%%%%%%%%%%%%%%%%%%%%%%%%%%%%%%%%
\beginSupplementaryMaterials

\appendix

\section{Detailed description of related work}
\label{sec:appendix1}

Tables \ref{table2a} and \ref{table2b} summarize and compare our approach with the most relevant work across seven dimensions. The first dimension is in terms of exploration approach to be token. CMAE \citep{CMAE} decides which dimensions to prioritize for exploration by calculating the entropy of each dimension of the state. EITI \citep{EITI} adopts a Value of Interaction (VoI) approach to quantify the influence of one agent's behavior on the expected returns of other agents. MASER adopts the maximum entropy principle to diversify the action choices. LAIES \citep{laies} specifies the external states of the environment and encourages the agent to explore them. FoX \citep{FoX} investigates the same formation in MADRL environments to enhance exploration. HMASD \citep{HMASD} enables agents to explore environments more efficiently based on the skills they learn. 
Different from these approaches, our method's exploration is based on the concept of influence scope introduced in this paper to delimit the exploration space for each agent. In this way, agents obtain count-based intrinsic rewards from the novel states discovered within their respective influence scope.

\begin{table}[!h]\small
	\centering{
		\begin{tabular}{l|l|l|l|l|l}
			\hline
			Methods & Exploration & \begin{tabular}[c]{@{}l@{}}Affiliation \\ of goal \end{tabular} & \begin{tabular}[c]{@{}l@{}}Individual \\ goal obtaining \end{tabular} & \begin{tabular}[c]{@{}l@{}}Assign credit \\ among agents \end{tabular} & \begin{tabular}[c]{@{}l@{}}Mutual  \\ Information \end{tabular} \\ \hline
			CM3    & ---   &  Individual   &   State            &  Implicit network        &   ---     \\
			CMAE    & Entropy        & Global         &   ---       &  Implicit network    &   ---     \\
			EITI    & VoI        & ---         &   ---       &  Counterfact VoI    &  Transitions   \\
			MASER   & Entropy        & Individual     &  Observation      &   Implicit  network   &  ---   \\
			ALMA    & ---        & Individual     &   Predefined     &    Implicit network           &   ---     \\
			LAIES   &  External state    &  ---  &   ---  &  Implicit  network  &  Transitions   \\
			FoX   & Formation          &  ---   &  ---   &   Implicit network    &  \begin{tabular}[c]{@{}l@{}}Formation\&  \\ latent variable\end{tabular} \\
			HMASD   & Skill          &  \begin{tabular}[c]{@{}l@{}}Global\&  \\ Individual\end{tabular}   &  latent variable   &   Implicit network    &  States\&skills   \\
			Ours    & \begin{tabular}[c]{@{}l@{}}Count within  \\ influence scope\end{tabular}   & \begin{tabular}[c]{@{}l@{}}Global\&  \\ Individual\end{tabular}  & \begin{tabular}[c]{@{}l@{}}Automatic  \\ decomposition\end{tabular}  &  \begin{tabular}[c]{@{}l@{}}Explicit if-else \\ function\end{tabular}  &  \begin{tabular}[c]{@{}l@{}}Elements of  \\ states\&actions\end{tabular} \\
			\hline
	\end{tabular}}
\vspace{0.2cm}
	\caption{Comparison to literature on mechanisms}
	\label{table2a}
\end{table}

\begin{table}[!h]
	\centering{
		\begin{tabular}{l|l|l}
			\hline
			Methods & \begin{tabular}[c]{@{}l@{}}No handcraft \\ knowledge \end{tabular} & \begin{tabular}[c]{@{}l@{}} Interpretable \\ credit assignment \end{tabular}  \\ \hline
			CM3    &   \usym{2714}   & \usym{2718}     \\
			CMAE   &   \usym{2714}   & \usym{2718}     \\
			EITI   &   \usym{2714}   & \usym{2718}     \\
			MASER  &   \usym{2714}   & \usym{2718}     \\
			ALMA   &   \usym{2718}   & \usym{2718}     \\
			LAIES  &   \usym{2718}   & \usym{2718}     \\
			FoX    &   \usym{2714}   & \usym{2718}     \\
			HMASD  &   \usym{2714}   & \usym{2718}     \\
			Ours   &   \usym{2714}   & \usym{2714}     \\
			\hline
	\end{tabular}}\vspace{0.2cm}
	\caption{Comparison to literature on usage of handcraft knowledge and interpretability}
	\label{table2b}
\end{table}

Table \ref{table2a} also compares our method with the previous methods from the perspective of the goals used in learning. `Affiliation of goal' in Table \ref{table2a} indicates whether the goal is designed for the global (team) or for the individual, and `individual goal obtaining' indicates how the representation of the individual goal is obtained. As shown in the table, our approach includes both global and individual goals. Our global goal is a state from the environment, while an individual goal is automatically decomposed from this global goal based on the influence scope of the individual agent.

As far as credit assignment is concerned, most previous approaches involving network policies have used a network to implicitly assign credit among agents through backpropagation of the gradient. One exception is EITI, which utilizes its proposed Value of Interaction (VoI) to exclude the contribution of other agents when evaluating individual behaviors. The credit assignment in our method is a explicit rule-based process. Our method determines whether the current actions of the agents impact a segment of the goal. If they do not, the reward for that segment will not be assigned to the agents.

Moreover, when it comes to the computation of mutual information, our approach considers the mutual information of each particular action of the agent on each dimension of the state value. This differs from calculating the mutual information of skill with respect to a hidden variable (e.g. HMASD), and from calculating the mutual information between transitions (e.g. EITI). As shown in Table \ref{table2b}, We also compare our method with previous approaches from two perspectives: the reliance on handcrafted knowledge and the interpretability of the credit assignment process. Unlike ALMA and LAIES, which rely on predefined goals or state variables, our method does not depend on handcrafted knowledge. Moreover, in contrast to prior methods, our approach provides interpretable credit assignment through its explicit if-else process.

\section{Detailed descriptions on experiments}
\label{sec:appendix2}

\subsection{Environments}

\begin{figure}[!t] 
	\centering 
    \begin{subfigure}[SMAC Example Task (8m)]{0.47\textwidth}
        \centering
        \includegraphics[width=\linewidth]{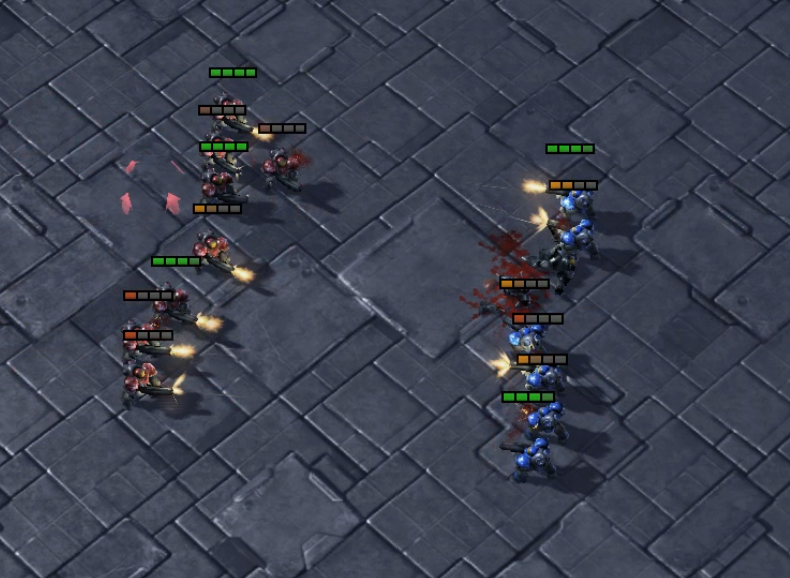}\label{fig:smac}
        \caption{}
    \end{subfigure}
    \begin{subfigure}[MPE Example Task (Navigation)]{0.4\textwidth}
        \centering
        \includegraphics[width=\linewidth]{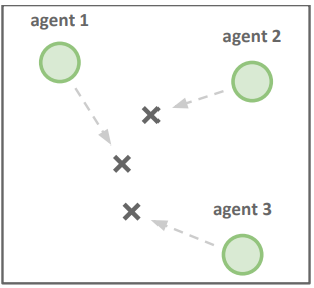}\label{fig:mpe}
        \caption{}
    \end{subfigure}
	\caption{Screenshots of environments.}
	\label{fig:smacmpe}
\end{figure}

\textbf{Environment.} We evaluate our method in two challenging sparse-reward benchmarks: (1) the starcraft multi-agent challenge (SMAC) \citep{SMAC}, and (2) the multiple-particle environment (MPE) \citep{MPE}, as shown in Figure \ref{fig:smacmpe}. In SMAC, agents need to learn cooperative policies to eliminate the enemies. SMAC has different reward setting as shown in Table \ref{tab:smac_reward}. In dense reward setting, agents intensively receive immediate reward for their actions, whereas in sparse and super sparse setting the rewards are delayed where the agents is rewarded only when rare events occur. In our experiments, we take the most challenging super sparse reward setting to validate the independence of our method from handcrafted rewards in sparse-reward tasks.

In addition to the SMAC environment, we also consider three tasks in MPE. In most SMAC scenarios, team rewards largely depends on state variables included by $D^c$ (e.g., enemy health) rather than state variables included by $D^{i-c}$ (e.g., individual agent coordinates). We consider MPE tasks to capture different patterns of dependency between team rewards and state variables.

\begin{itemize}
    \item \textbf{Navigation.} In this task, agents are required to occupy different landmarks, and the team receives a reward of $+1$ whenever a landmark is occupied. The state information includes the positions of all agents and landmarks. This is a typical scenario in which the reward depends solely on $D^{i-c}$, as there is no inherent joint influence ($D^c = \emptyset$). We use this task to evaluate the performance of ISA in environments where rewards rely only on individual-level contributions.
    \item \textbf{Shooting.} In this task, agents must move to specific positions and collectively shoot a target multiple times. The team will receive a reward of $+1$ only when both moving and shooting are completed. The state contains the number of times the target has been shot, which is jointly influenced by all agents ($D^c$), as well as the positions of individual agents ($D^{i-c}$). The rewards of this task therefore involves both $D^c$ and $D^{i-c}$, and is used to validate ISA in environments where team rewards depend on the combination of joint influence and individual contributions.
    \item \textbf{Unlock.} In this task, each agent holds a different key that must be used to unlock the corresponding locks, and the team receives a reward of $+1$ whenever a lock is successfully opened. Unlock is an environment similar to Navigation, but with a stricter condition. In Navigation, an agent can occupy any landmark to obtain a reward. In contrast, in Unlock an agent only receives a reward by opening its corresponding lock. We use this environment to evaluate ISA under stricter conditions.
\end{itemize}

Moreover, the state space in SMAC is continuous, while that in MPE is discrete. Detailed specifications of both environments are summarized in Table \ref{tab:env}.

\begin{table}[!h] \small
	\centering 
	\begin{tabular}{c|cccc}
		\hline
		& number of states & number of actions & number of agents & dimension of state \\ \hline  
		3m          &  $+\infty$                & 9                 & 3                & 46                 \\
		2s\_vs\_1sc &  $+\infty$                & 7                 & 2                & 25                 \\
		8m          &  $+\infty$                & 14                & 8                & 166                \\
		Navigation  &  $20^{14}$                & 5                 & 7                & 14                 \\
		Shooting    &  $20^6 \times 50$                & 6                 & 3                & 7                  \\
		Unlock      &  $20^{14}$                & 5                 & 7                & 14                 \\ \hline
	\end{tabular}\vspace{0.2cm}
	\caption{Environmental information for tasks in SMAC and MPE}
	\label{tab:env}
\end{table}

\subsection{Baselines}
The detailed information of the baselines are shown as follows.

\begin{itemize}
    \item \textbf{Classical MADRL methods.} QMIX, COMA, and IPPO are 3 classical  MADRL algorithms. QMIX using a mixing network to assign credit among agents. COMA trains a counterfactual value function for credit assignment. IPPO directly uses team reward as individual rewards. CMAE is an exploration method designed for sparse-reward multi-agent tasks.
    \item \textbf{Similar method with ours.} MASER, like our method, employs a distance-based intrinsic reward measure for individual goal achievement, but the representation of individual goals is encoded based on observations.
    \item \textbf{State-of-the-art methods.} FoX is an state-of-art exploration MADRL method in terms of performance in tasks with few immediate rewards. HMASD explores and learns the tasks with skills, which is the state-of-art method in terms of performance in tasks with super sparse setting of SMAC.
\end{itemize}

For the implement of MASER, FoX, and HMASD, we use the code released by the authors. For the implement of classical  MADRL algorithms QMIX, COMA, and IPPO, we employ a widely code framework released by \citep{hu2021rethinking}. Because the code provided by authors of CMAE only work for 2 agents environments and the do not provide code for SMAC domain, we use the performance data reported in their paper.

\subsection{Hyperparameters}

\begin{table}[!h]
	\centering
	\begin{tabular}{l|l}
		\hline
		Hyper-parameter & Description \\ \hline
		$\delta$ & Threshold in Equation (\ref{Equ:I}) \\
		$\lambda$ & Scale factor for Hamming distance \\
		$\alpha_1, \beta_1$ & Scale factors in Equation (\ref{Equ:GCR}) and (\ref{Equ:ER}) \\
		$\alpha_2, \beta_2$ & Scale factors for environment reward \\
		$N$ & Number of transitions for calculate  Equation (\ref{Equ:I}) \\
		$L$ & Goal buffer Length in Alg. \ref{alg1} \\
		$w_{bin}$ & Width of equal binning to discretize $\Delta s^k$ \\
		${l}_{hidden}$ & Size of hidden layer \\
		$l_{buffer}$ & Size of episode buffer for policy training \\
		$n_{running}$ & Number of processes for data collection \\
		$n_{training}$ & Batch size for policy training \\
		$lr$ & Learning rate \\
		$\lambda_{critic}$ & Scale factor for updating critic \\
		$\lambda_{entropy}$ & Scale factor for entropy regularization \\
		$\lambda_{gae}$ & GAE factor \\
		$c$ & Clipping value to restrict the update \\
		\hline
	\end{tabular}
	\caption{Descriptions of hyperparameters used in ISA}
	\label{tab:hyper_a}
\end{table}

\begin{table}[!h]\small
	\centering
	\begin{tabular}{l|llllllll}
		\hline
		& 3m & 2s\_vs\_1sc & 8m & 8m\_vs\_9m & 10m\_vs\_11m & Navigation & Shooting & Unlock \\ \hline
		$\delta$ & 0.3 & 0.3 & 0.3 & 0.3 & 0.3 & 0.3 & 0.3 & 0.3 \\
		$\lambda$ & 50 & 50 & 10 & 10 & 10 & 0 & 0 & 0 \\
		$\alpha_1, \beta_1$ & 0 & 0 & 0 & 0 & 0 & 0.2 & 0.2 & 0.2 \\
		$\alpha_2, \beta_2$ & 0 & 0 & 10 & 10 & 10 & 1 & 1 & 1 \\
		$N$ & 2,000 & 2,000 & 10,000 & 10,000 & 10,000 & 2,000 & 2,000 & 2,000 \\
		$L$ & 1 & 1 & 1 & 1 & 1 & 1 & 1 & 1 \\
		$w_{bin}$ & 0.01 & 0.05 & 0.01 & 0.01 & 0.01 & 0.01 & 0.01 & 0.01 \\
		$l_{hidden}$ & 64 & 64 & 64 & 64 & 64 & 12 & 12 & 12 \\
		$l_{buffer}$ & 64 & 64 & 64 & 64 & 64 & 32 & 32 & 32 \\
		$n_{running}$ & 8 & 8 & 8 & 8 & 8 & 8 & 8 & 8 \\
		$n_{training}$ & 64 & 64 & 64 & 64 & 64 & 32 & 32 & 32 \\
		$lr$ & 0.0005 & 0.0005 & 0.0002 & 0.0002 & 0.0002 & 0.0005 & 0.0005 & 0.0005 \\
		$\lambda_{critic}$ & 0.5 & 0.5 & 0.5 & 0.5 & 0.5 & 0.5 & 0.5 & 0.5 \\
		$\lambda_{entropy}$ & 0.001 & 0.001 & 0.001 & 0.001 & 0.001 & 0.001 & 0.001 & 0.001 \\
		$\lambda_{gae}$ & 0.95 & 0.95 & 0.95 & 0.95 & 0.95 & 0.95 & 0.95 & 0.95 \\
		$c$ & 0.2 & 0.2 & 0.2 & 0.2 & 0.2 & 0.2 & 0.2 & 0.2 \\
		\hline
	\end{tabular}\vspace{0.2cm}
	\caption{Task-specific hyperparameters used in ISA}
	\label{tab:hyper_b}
\end{table} 

We show the description of our hyperparameters in Table \ref{tab:hyper_a} and their detailed settings for different tasks in Table \ref{tab:hyper_b}. The hyperparameter settings for the learning part of ISA make reference to IPPO \citep{IPPO}. During the development of ISA, we tried different hyperparameters. As elaborated in Remark 1, too large or too small values for threshold $\delta$ are not inappropriate to capture the inherent influence scope of agent in the environment. In the development of ISA, we tried values of $\delta$ between 0 and 0.6. The results show that the inherent influence scope of agents can be accurately captured when $\delta \in [0.15, 0.45]$ in the 3m task. We finally chose $\delta=0.3$ and applied it to all other tasks. The scale factor for Hamming distance is used to scale the intrinsic reward whether or not each value of attribute on current state achieving the value in the goals. During the development, we tried the value of $\lambda$ from $\{0, 10, 50\}$ in SMAC domain. For factors $\alpha_2, \beta_2$ that scale the environmental reward, we tried their value from $\{0, 10, 50\}$ in the 8m task. In addition, because there are more agents and more individual actions in 8m, we took a larger $N=10000$ in this task to collect more data to evaluate the mutual information. Besides, we tried larger $w_{bin} = 0.05$ on 2s\_vs\_1sc to ignore the minor changes in the state's attributes, and tried smaller $lr = 0.0002$ on 8m to stabilize the learning. Apart from these mentioned tries, the other hyperparameters always adopt the values in Table \ref{tab:hyper_b} during the development of ISA.

\end{document}